\documentclass[10pt,twocolumn,letterpaper]{article}

\usepackage[pagenumbers]{cvpr} 

\usepackage{graphicx}
\usepackage{amsmath}
\usepackage{amssymb}
\usepackage{booktabs}

\usepackage[pagebackref,breaklinks,colorlinks=true]{hyperref}

\usepackage[capitalize]{cleveref}
\crefname{section}{Sec.}{Secs.}
\Crefname{section}{Section}{Sections}
\Crefname{table}{Table}{Tables}
\crefname{table}{Tab.}{Tabs.}

\def\cvprPaperID{4147} 
\def\confName{CVPR}
\def\confYear{2022}

\usepackage{overpic}
\usepackage{enumitem} 
\usepackage{overpic} 
\usepackage{color}
\usepackage{microtype} 

\definecolor{turquoise}{cmyk}{0.65,0,0.1,0.3}
\definecolor{purple}{rgb}{0.65,0,0.65}
\definecolor{dark_green}{rgb}{0, 0.5, 0}
\definecolor{orange}{rgb}{0.8, 0.6, 0.2}
\definecolor{red}{rgb}{0.8, 0.2, 0.2}
\definecolor{darkred}{rgb}{0.6, 0.1, 0.05}
\definecolor{blueish}{rgb}{0.0, 0.3, .6}
\definecolor{light_gray}{rgb}{0.7, 0.7, .7}
\definecolor{pink}{rgb}{1, 0, 1}
\definecolor{greyblue}{rgb}{0.25, 0.25, 1}

\newcommand{\loss}[1]{\mathcal{L}_\text{#1}}
\newcommand{\expect}{\mathbb{E}}
\newcommand{\real}{\mathbb{R}}
\newcommand{\calds}{{\cal D}_{\cal S}}
\newcommand{\caldt}{{\cal D}_{\cal T}}
\newcommand{\caldfs}{{\tilde {\cal D}}_s}
\newcommand{\caldft}{{\tilde {\cal D}}_t}

\newcommand{\Fig}[1]{Fig.~\ref{fig:#1}}

\newcommand{\Table}[1]{Table~\ref{tab:#1}}

\newcommand{\Equation}[1]{Equation~\ref{eq:#1}}
\newcommand{\Sec}[1]{Sec.~\ref{sec:#1}}

\usepackage{amsthm}
\newcommand{\Def}[1]{Definition~\ref{def:#1}}
\newcommand{\Prop}[1]{Proposition~\ref{prop:#1}}
\newcommand{\Lemma}[1]{Lemma~\ref{lemma:#1}}
\newcommand{\Thm}[1]{Theorem~\ref{thm:#1}}

\usepackage{blindtext}

\renewcommand{\paragraph}[1]{\vspace{1em}\noindent\textbf{#1}.}

\usepackage[normalem]{ulem}
\useunder{\uline}{\ul}{}
\usepackage[tabel,xcdraw]{xcolor}
\newtheorem{lemma}{Lemma}
\newtheorem{thm}{Theorem}
\newtheorem{definition}{Definition}
\newtheorem{prop}{Proposition}

\begin{document}

\title{Reusing the Task-specific Classifier as a Discriminator:\\
Discriminator-free Adversarial Domain Adaptation}

\author{Lin Chen\thanks{indicates equal contribution.}\,\quad Huaian Chen\footnotemark[1]\,\quad Zhixiang Wei\quad Xin Jin\quad Xiao Tan\quad Yi Jin\thanks{Corresponding author.}\,\quad Enhong Chen\\
University of Science and Technology of China\\
{\tt\small \{chlin, anchen, zhixiangwei, jinxustc, tx2015\}@mail.ustc.edu.cn}\\
{\tt\small \{jinyi08, cheneh\}@ustc.edu.cn}}
\maketitle
\begin{abstract}
 Adversarial learning has achieved remarkable performances for unsupervised domain adaptation (UDA). Existing adversarial UDA methods typically adopt an additional discriminator to play the min-max game with a feature extractor. However, most of these methods failed to effectively leverage the predicted discriminative information, and thus cause mode collapse for generator. In this work, we address this problem from a different perspective and design a simple yet effective adversarial paradigm in the form of a discriminator-free adversarial learning network (DALN), wherein the category classifier is reused as a discriminator, which achieves explicit domain alignment and category distinguishment through a unified objective, enabling the DALN to leverage the predicted discriminative information for sufficient feature alignment. Basically, we introduce a Nuclear-norm Wasserstein discrepancy (NWD) that has definite guidance meaning for performing discrimination. Such NWD can be coupled with the classifier to serve as a discriminator satisfying the K-Lipschitz constraint without the requirements of additional weight clipping or gradient penalty strategy. Without bells and whistles,  DALN compares favorably against the existing state-of-the-art (SOTA) methods on a variety of public datasets. Moreover, as a plug-and-play technique, NWD can be directly used as a generic regularizer to benefit existing UDA algorithms. Code is available at \url{https://github.com/xiaoachen98/DALN}.
\end{abstract}
\section{Introduction}
\label{sec:intro}
\begin{figure}[htbp]
\centering
\subfloat[Bi-classifier \quad  \quad \, \, (b) Extra Discriminator \qquad \qquad (c) Ours \quad \quad]{\includegraphics[width=\linewidth]{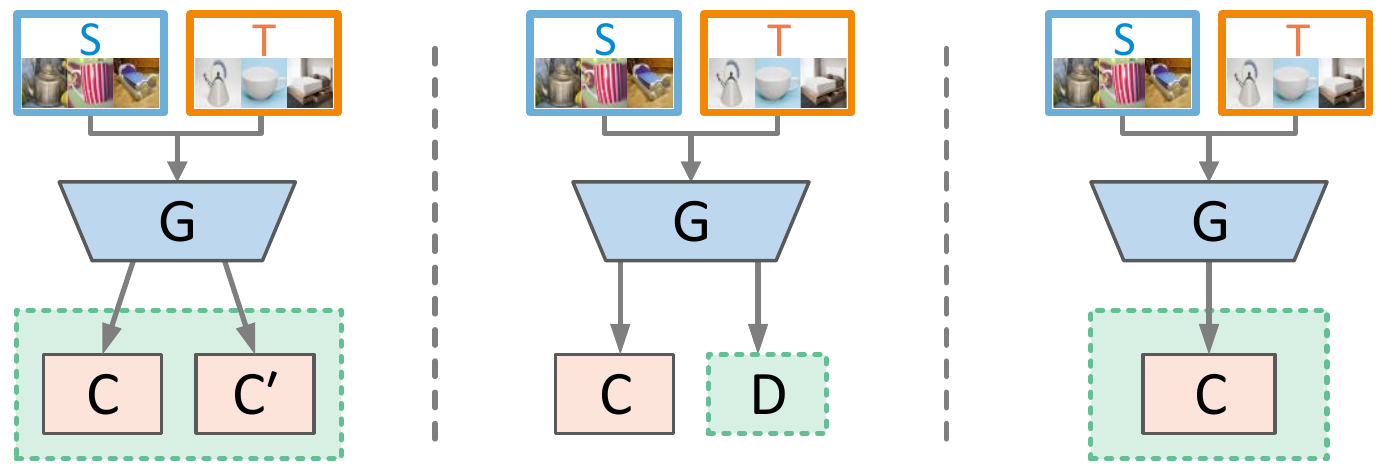}}
\caption{
Illustration of different adversarial paradigms, in which $G$, $C$, and $D$ denote the feature extractor, task-specific classifier, and discriminator, respectively. Different from typical paradigms that adopt an (a) additional classifier $C'$ (called bi-classifier) or (b) additional discriminator $D$, we present a different perspective for UDA and introduce a simple but effective adversarial paradigm illustrated in (c), in which the original task-specific classifier $C$ is reused as a implicit discriminator, achieving explicit domain alignment and category distinguishment via a unified objective.
}
\label{fig:teaser}
\end{figure}
Deep neural networks (DNNs) have achieved a significant progress in many computer vision tasks \cite{he2016deep,chen2017deeplab,ren2015faster,carion2020end}. However, the success of these methods highly depends on large amounts of annotated data \cite{wang2018deep,zhao2020review,glorot2011domain}, which is extremely time-consuming and expensive to obtain. Moreover, due to the discrepancy \cite{pan2009survey,pan2010domain} between training data and real-world testing data, the DNN model trained on annotated  data may suffer from a dramatic performance decline in testing set despite  extensive annotation efforts. To address this problem, unsupervised domain adaptation (UDA) \cite{chen2019transferability,cui2020gradually,wei2021metaalign,na2021fixbi}, which aims to transfer knowledge from a labeled source domain to an unlabeled target domain in the presence of a domain shift, has been deeply explored.

Inspired by the theoretical analysis of Ben-David \etal \cite{ben2007analysis}, the existing UDA methods usually explore the idea of learning domain-invariant feature representations. Generally, these methods can be categorized into two branches, i.e., moment matching methods \cite{tzeng2014deep,long2015learning,zhang2019bridging,li2020enhanced,NEURIPS2018_ab88b157} and adversarial learning methods \cite{ganin2016domain,saito2018maximum,gao2021gradient,NEURIPS2018_ab88b157}. Moment matching methods explicitly reduce the domain shift by matching a well-defined distribution discrepancy of the source and target domain features. Adversarial learning methods implicitly mitigate the domain shift by playing an adversarial min-max two-player game, which drives the generator to extract indistinguishable features to fool the discriminator. Encouraged by the remarkable performance achieved by adversarial learning, increasingly more researchers have been devoted to developing a UDA method based on 
an adversarial paradigm \cite{tang2020discriminative,du2021cross,lu2020stochastic,cui2020gradually,li2021semantic,Li21BCDM}.

Basically, adversarial learning-based UDA methods usually follow two lines of adversarial paradigms.
One line\cite{saito2018maximum,lu2020stochastic,Li21BCDM,du2021cross,lee2019sliced} leverages the disparity of two task-specific classifiers $C$ and $C'$ (as shown in \Fig{teaser}(a)), which can be deemed as a discriminator, to implicitly achieve adversarial learning and improve feature transferability. This paradigm enables UDA methods to reduce the class-level domain discrepancy. However, the methods following this paradigm are prone to be affected by ambiguous predictions and thus hinder the adaption optimization. 
The other line \cite{NEURIPS2018_ab88b157, ganin2016domain, gao2021gradient, cui2020gradually} directly constructs an additional domain discriminator $D$ as shown in \Fig{teaser}(b), which improves the feature transferability by sufficiently confusing the cross-domain feature representations. However, the methods following this paradigm usually focus on the domain-level feature confusion, which may hurt the category-level information and thus cause mode collapse problem \cite{tang2020discriminative,kurmi2019looking}.

To address these problems, we present a different perspective for UDA and introduce a simple but effective adversarial paradigm illustrated in \Fig{teaser}(c). In this paradigm, the original task-specific classifier is coupled with a novel discrepancy to serve as a discriminator/critic, which simultaneously achieves  domain alignment and category distinguishment through a unified objective, enabling the model to leverage the predicted discriminative information to capture the multi-modal structures \cite{NEURIPS2018_ab88b157,gao2021gradient} of the feature distributions. Particularly, when classifier $C$ is used for classification, it helps achieve category-level distinguishment; furthermore, when $C$ serves as a discriminator, it achieves feature-level alignment. The novel discrepancy, called Nuclear-norm Wasserstein discrepancy (NWD), leverages the advantages of the Nuclear norm and 1-Wasserstein distance to encourage the prediction determinacy and diversity. Different from the discrepancy metrics used in existing adversarial methods \cite{ganin2016domain,tang2020discriminative,zhang2019domain}, the NWD not only has a promising theoretical generalization bound but also has definite guidance meaning for performing discrimination, i.e., naturally giving high scores to the source domain samples and low scores to the target domain samples due to the supervised training on the source domain. Such guidance encourages the intra-class and inter-class correlations of the target domain to approach those of the source domain. Moreover, in contrast to the existing Wasserstein discrepancy used in recent work \cite{shen2018wasserstein}, the NWD enables the adversarial UDA paradigm to satisfy the K-Lipschitz constraint without the need to set up an additional weight clipping \cite{arjovsky2017wasserstein} or gradient penalty \cite{10.5555/3295222.3295327}.

Based on the introduced paradigm, we propose a discriminator-free adversarial learning network (DALN), which achieves adversarial UDA classification without explicit domain discriminator. Benefiting from the definite guidance of the NWD, the proposed DALN converges rapidly and achieves competitive prediction determinacy and diversity. Note that, the DALN is considerably different from recent approaches \cite{tang2020discriminative,zhang2019domain} that integrate the discriminator into the classifier. DALN directly reuses the original task-specific classifier without requiring any additional components, making it quite simple and efficient. Extensive experiments on a variety of datasets demonstrate that the proposed DALN outperforms the existing state-of-the-art (SOTA) methods. Moreover, we show that the proposed NWD is general and plug-and-play, which can be used as a regularizer to benefit the existing methods, which helps them achieve more competitive performance. The main contributions of this work are summarized as follows:

\begin{itemize}[leftmargin=*]
\setlength\itemsep{-.3em}
\item We present a different perspective for UDA by introducing a simple yet effective adversarial paradigm, in which the original task-specific classifier is reused as a discriminator. Based on this, we propose a new UDA method, namely DALN, which can leverage the predicted discriminative information for sufficient feature alignment.

\item We introduce a new discrepancy, termed NWD, which has a theoretical generalization bound and definite guidance meaning. Such discrepancy enables the implicitly constructed discriminator to satisfy the K-Lipschitz constraint without  the requirements of additional weight clipping and gradient penalty strategies. 

\item Without bells and whistles but only a few lines of code, the proposed method achieves highly competitive performance on various public datasets. By taking the proposed NWD as a regularizer for existing methods, these methods can achieve more competitive performance.
\end{itemize}
\begin{figure*}
\begin{center}
\includegraphics[width=\textwidth]{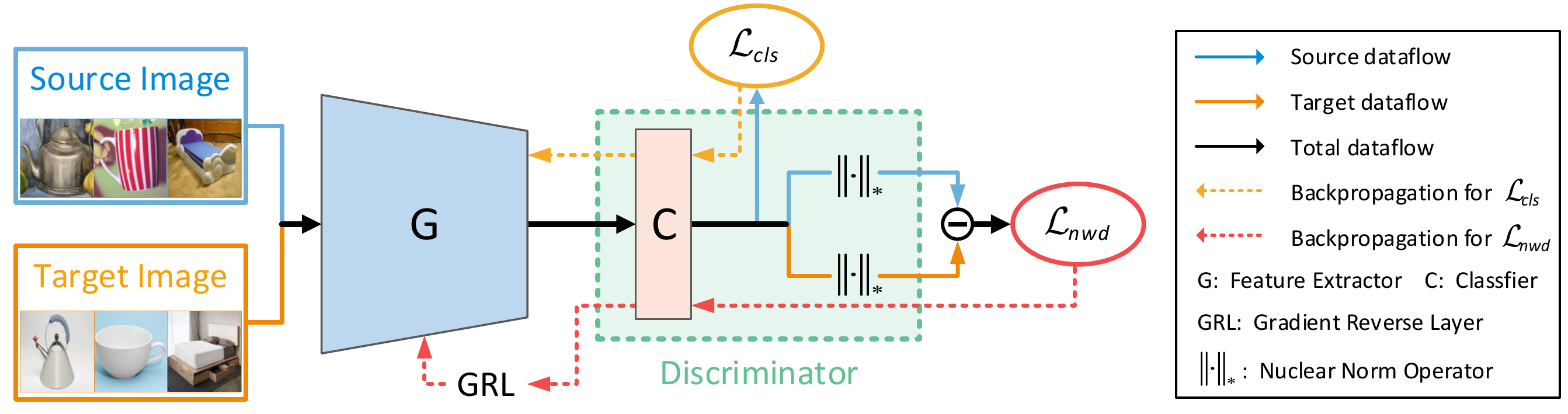}
\end{center}
\caption{
An overview of the adversarial paradigm in the form of DALN, which consists of a feature extractor $G$ and a task-specific classifier $C$. ${\cal L}_{cls}$ is used to guarantee a low source risk for the source domain, and ${\cal L}_{nwd}$ is used to empirically estimate the NWD that can be coupled with classifier $C$ to implicitly serve as a discriminator. The gradient reverse layer is used to help perform the adversarial learning. 
}
\label{fig:overview}
\end{figure*} 

\section{Related Works}
\label{sec:related}

The existing UDA methods can be mainly divided into two categories, i.e., moment matching methods \cite{tzeng2014deep,long2015learning,long2017deep,zhang2019bridging} and adversarial learning methods \cite{ganin2016domain,NEURIPS2018_ab88b157,zhang2019domain,saito2018maximum,du2021cross}.

\noindent\textbf{Moment Matching Methods.} Moment matching methods learn domain-invariant feature representations by matching a well-defined moment-based distribution discrepancy \cite{zhao2020review} across domains. Typically, DDC \cite{tzeng2014deep} attempted to explicitly align the learned feature distributions across domains by minimizing the maximum mean discrepancy (MMD). Later, methods in \cite{long2015learning,long2017deep} improved DDC by performing alignment with multi-kernel maximum mean discrepancy (MK-MMD) and joint maximum mean discrepancy (JMMD), respectively. In addition, MDD \cite{zhang2019bridging} proposed margin disparity discrepancy (MDD) to reduce the distribution discrepancy.

\noindent\textbf{Adversarial Learning Methods.} Inspired by generative adversarial network (GAN) \cite{goodfellow2014generative}, adversarial learning methods learn domain-invariant features via a  min-max two-player game. As one of the earliest attempts, DANN \cite{ganin2016domain} introduced an additional discriminator to distinguish the features generated by the feature extractor, which successfully achieves the domain-level adaptation. The success of the DANN exhibits the ability to improve UDA with the GAN model. Later, FGDA \cite{gao2021gradient} leveraged a discriminator to distinguish the gradient distribution of features, which achieved better performance for reducing domain discrepancy. Inspired by the conditional GAN \cite{mirza2014conditional}, methods in \cite{NEURIPS2018_ab88b157,pei2018multi} combined the predicted discriminative information with learned features to improve feature alignment. Additionally, DADA \cite{tang2020discriminative} attempted to couple the task-specific classifier with the domain discriminator to align the joint distributions of two domains. Although these methods successfully learn domain-invariant features, they cannot guarantee an appropriate divergence used for the discriminator when the support sets of two distributions do not overlap with each other \cite{arjovsky2017wasserstein}.

In addition to the methods adopting an additional discriminator, some studies attempted to use two task-specific classifiers (called bi-classifier), in which the disparity of two task-specific classifiers can be deemed as a discriminator \cite{lee2019sliced,du2021cross,zhang2019domain,saito2018maximum,lu2020stochastic}, to implicitly achieve the adversarial learning. Representatively, MCD \cite{saito2018maximum} simply used the L1 distance to measure the intra-class divergence of two classifiers. SWD \cite{lee2019sliced} proposed using sliced Wasserstein discrepancy instead of L1 distance to obtain a more geometrically meaningful intra-class divergence. CGDM \cite{du2021cross} additionally introduced the cross-domain gradient discrepancy to further alleviate the domain discrepancy. Although these methods have achieved considerable improvements in reducing domain discrepancy, most of them consider only the intra-class divergence between predictions, which may result in ambiguous predictions.

Different from the aforementioned methods adopting an additional discriminator or classifier, we reuse the original task-specific classifier by coupling it with the designed NWD, implicitly constructing a discriminator/critic satisfying the K-Lipschitz constraint without the requirements of additional weight clipping or gradient penalty strategy.
\section{Method}
\subsection{Recap of Preliminary Knowledge}\label{sec:3.1}
Given a labeled source domain set $\left\{ {\left( {x_i^s,y_i^s} \right)} \right\}_{i = 1}^{{N^s}}$ with ${N_s}$ samples drawn from source domain ${{\cal D}_{\cal S}}$, where $x_i^s \in {{\cal X}_s}$, $y_i^s \in {{\cal Y}_s}$, and label ${y^s}$ covers $k$ classes, and an unlabeled domain target set $\left\{ {x_i^t} \right\}_{i = 1}^{{N^t}}$  with ${N_t}$ samples drawn from target domain ${{\cal D}_{\cal T}}$, where $x_i^t \in {{\cal X}_t}$, the goal of this work is to learn a deep UDA model for learning domain-invariant representations and achieving reliable predictions on the target domain. This model consists of a feature generator $G\left(  \cdot  \right)$ that maps the input data to the features $f \in {\mathbb{R}^d}$, i.e., ${f^s} = G\left( {{x^s}} \right)$ and ${f^t} = G\left( {{x^t}} \right)$,  and a task-specific classifier $C\left(  \cdot  \right)$ that generates corresponding predictions $p \in {\mathbb{R}^k}$, i.e., ${p^s} = C\left( {{f^s}} \right)$ and ${p^t} = C\left( {{f^t}} \right)$. To this end, the existing adversarial UDA approaches usually take an additional discriminator or classifier. Typically, many popular methods \cite{ganin2016domain,NEURIPS2018_ab88b157} use an additional discriminator $D\left(  \cdot  \right)$ to achieve adversarial UDA by optimizing object classification loss ${\cal L}_{cls}$ and domain adversarial loss ${\cal L}_{adv}$: 
\begin{align}
    {\cal L}_{cls}  = {} & {\expect_{\left( {x_i^s,y_i^s} \right) \sim \calds}}{\loss{ce}\left( {C\left( G\left( {x_i^s} \right)\right),y_i^s} \right)}, \label{eq:1}\\
{\cal L}_{adv} = {} & {\expect_{G\left( {x_i^s} \right) \sim {\tilde {\cal D}}_s}}\log \left[ {D\left( {G\left( {x_i^s} \right)} \right)} \right] \notag \\
& + {\expect_{G\left( {x_i^t} \right) \sim {{\tilde {\cal D}}_t}}}\log \left[ {1 - D\left( {G\left( {x_i^t} \right)} \right)} \right], \label{eq:2}
\end{align}
where $\caldfs$ and $\caldft$ denote the induced feature distributions of $\calds$ and $\caldt$, respectively, and $\loss{ce} \left( \cdot, \cdot \right)$ is  the cross-entropy loss function. However, we find that the original task-specific classifier $C$ has an implicit discriminative ability for the source domain and target domain, and can be directly used as a discriminator (see \Sec{3.2}). Inspired by this observation, as shown in Fig. \ref{fig:overview}, we propose a simple yet effective adversarial paradigm for adversarial UDA: reusing the task-specific classifier as a discriminator.

\subsection{Reusing the Classifier as a Discriminator}\label{sec:3.2}
\begin{figure}[htbp]
\centering
\vspace{-2mm}
\subfloat[Source domain]{\includegraphics[width=0.49\linewidth]{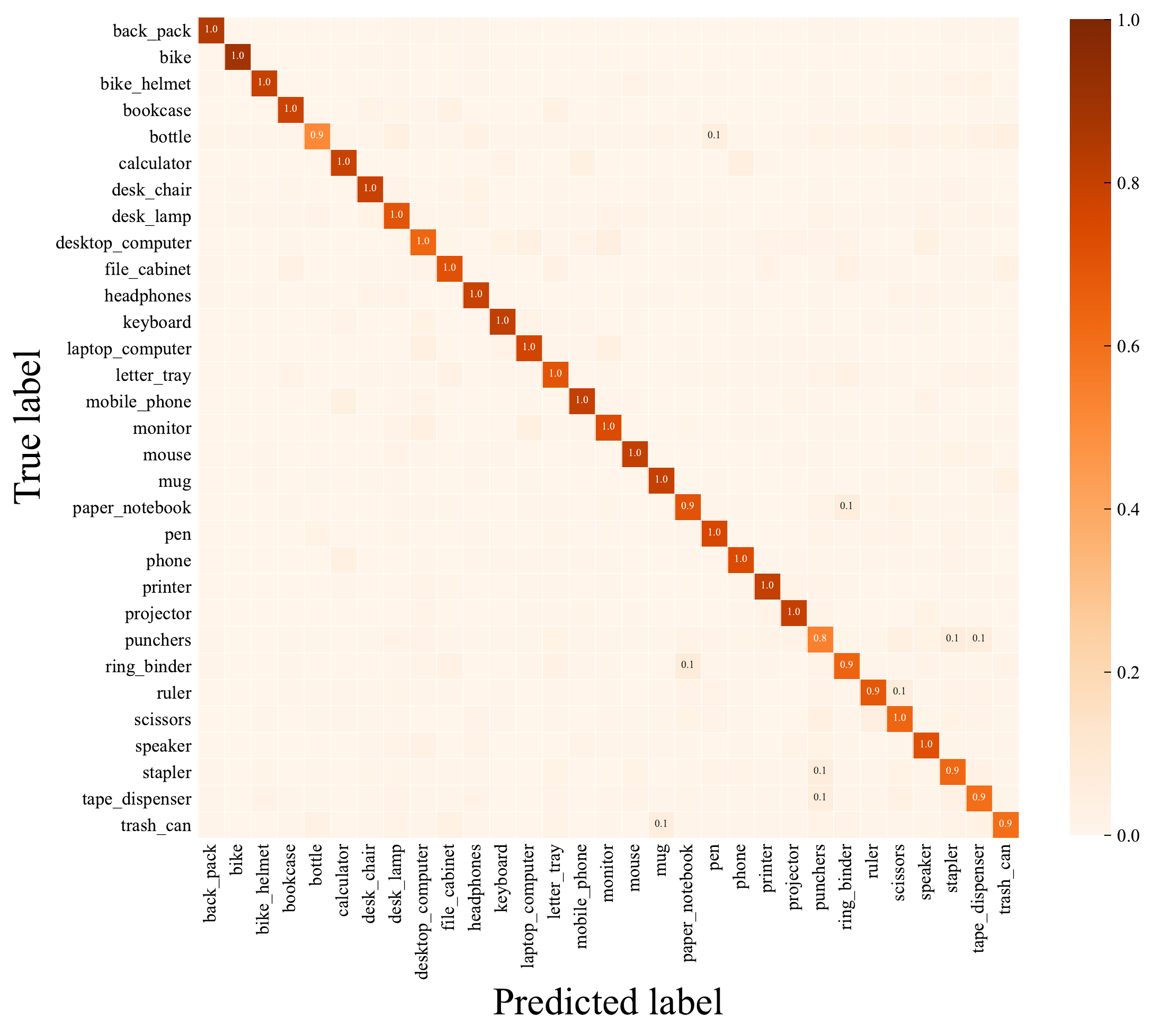}}
\subfloat[Target domain]{\includegraphics[width=0.49\linewidth]{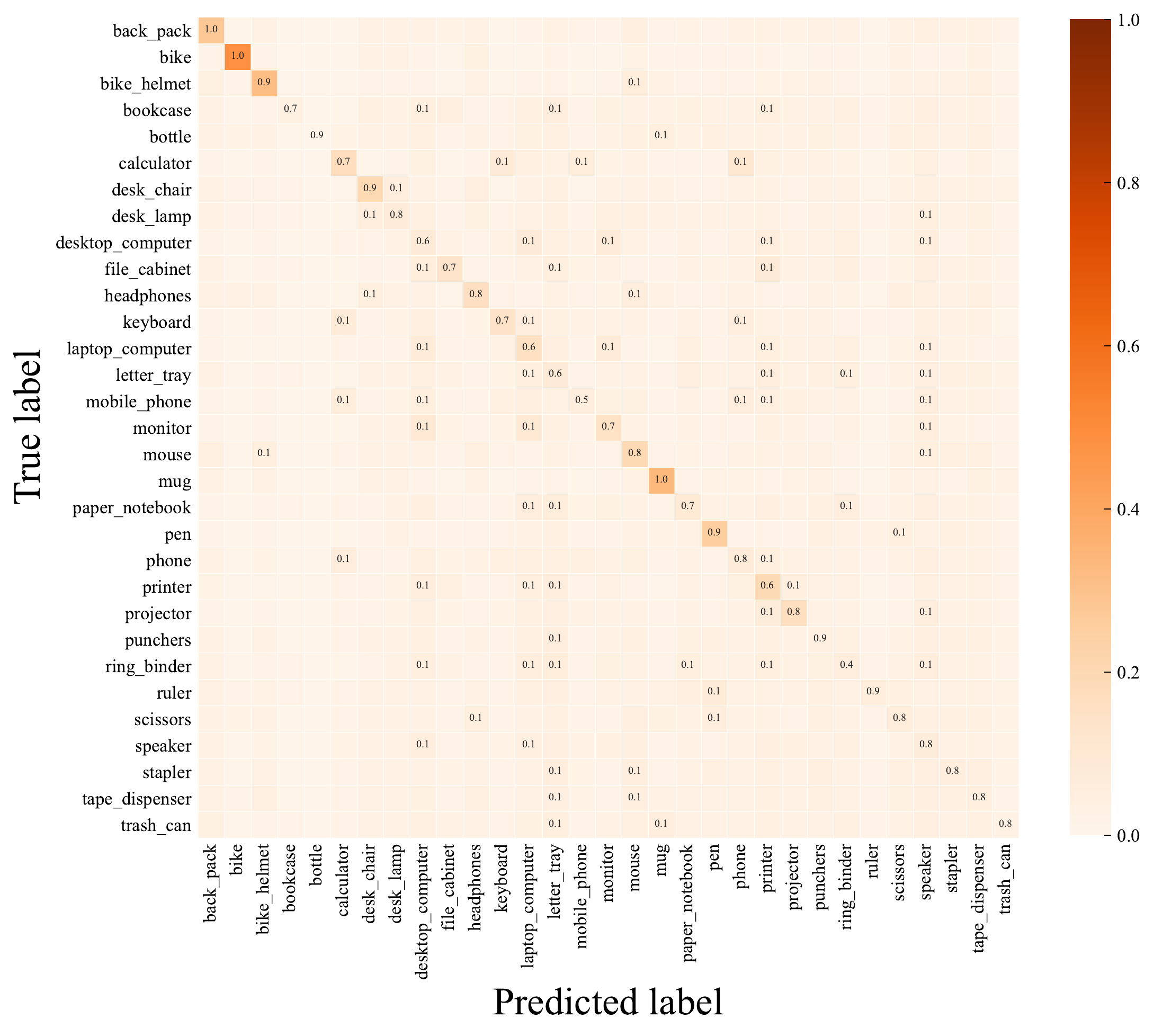}}
\vspace{-1mm}
\caption{
The self-correlation matrices of the predictions on the source and target domains based on a DNN model trained only with the source domain data on task A→W of Office-31. (Zoom in for a clear visualization.)
}
\label{fig:correlation}
\end{figure}

\noindent\textbf{Motivation Re-clarification.} As we claimed before, the original task-specific classifier has an implicit discriminative  ability for the source domain and the target domain. Fig. \ref{fig:correlation} presents the self-correlation matrices of the predictions on the source and target domains based on a model trained with the source-only data. For the source domain, benefiting from the supervised training, the values of the self-correlation matrix are concentrated on the main diagonal. In contrast, for the target domain, the prediction generates larger values on the off-diagonal elements due to the lack of supervision. Therefore, the intra-class and inter-class correlations represented in the self-correlation matrix are capable of constructing the adversarial critic.

\noindent\textbf{Rethinking the Intra-class and Inter-class Correlations.} Given a prediction matrix $Z \in {\real ^{b \times k}}$  predicted by $C$ that contains the prediction probabilities of $k$ categories multiplied by $b$ samples, the self-correlation matrix $R \in \real ^{k \times k}$ can be calculated by  $R = Z ^ T Z$, where the prediction matrix $Z=C\left(f\right)$ satisfies
\begin{align}\label{eq:4}
    \sum\limits_{j = 1}^k & {{Z_{i,j}} = 1}  \quad \forall i \in 1 \ldots b \notag \\
    & {{Z_{i,j}} \ge 0}  \quad \forall i \in i \ldots b,j \in 1 \ldots k.
\end{align}
For a self-correlation matrix $R$, the main diagonal elements represent the intra-class correlation and the off-diagonal elements denote the inter-class correlation or confusion \cite{jin2020minimum}. For convenient, in this work, we define the overall intra-class correlation  as $I_a$  and the overall inter-class correlation as $I_e$:
\begin{align}\label{eq:5}
    {I_a} = \sum\limits_{i,j = 1}^k {{R_{ij}}} & \quad {I_e} = \sum\limits_{i \ne j}^k {{R_{ij}}}.
\end{align}
For the source domain, the prediction contributes to a large $I_a$ and a small $I_e$; while for the target domain, the prediction generally produces a relatively small $I_a$ and large $I_e$ due to the lack of supervised training. Thus, $I_a - I_e$ can be used to represent the domain discrepancy. According to \Equation{4}, $I_a$ and $I_e$ satisfy ${I_a} + {I_e} = b$. Meanwhile, $I_a$ is equal to the Frobenius norm of prediction matrix $Z$, i.e, ${I_a} = {\left\| Z \right\|_F}$.  Thus, we have ${I_a} - {I_e} = 2{\left\| Z \right\|_F} - b$. $Z$ is predicted via the classifier $C$, so we can use $2{\left\| {C} \right\|_F} - b$ as a correlation critic function, which naturally gives high scores for the source domain samples and low scores for the target domain samples due to the supervised training on source domain. Moreover, considering weight 2 and bias b are both constants, the ${\left\| {C} \right\|_F}$ can be directly used as a correlation critic function.

\noindent\textbf{From Correlations Critic to 1-Wasserstein Distance.} Inspired by the WGAN \cite{arjovsky2017wasserstein}, a straightforward idea is to introduce an additional discriminator $D$ to learn a K-Lipschitz critic function $h$ expected to give high scores to source representations $f \in \caldfs$  and low scores to the target representations $f \in \caldft$, and measure the 1-Wasserstein distance ${W_1}\left( \caldfs, \caldft \right)$ between two feature distributions ${{\tilde {\cal D}}_s}$, $\caldft$ by
\begin{equation}\label{eq:6}
    {W_1}\left( \caldfs, \caldft \right) = \mathop {\sup }\limits_{{{\left\| h \right\|}_L} \le K} {\expect_{f \sim {\caldfs}}}\left[ {h\left( f \right)} \right] - {\expect_{f \sim {\caldft}}}\left[ {h\left( f \right)} \right],
\end{equation}
where ${\left\|  \cdot  \right\|_L}$ denotes the Lipschitz semi-norm \cite{villani2009optimal}, and $K$ denotes the Lipschitz constant. But, as we claimed above,  ${\left\| {C} \right\|_F}$  has exactly definite critic meaning to serve as $D$. Then, the domain discrepancy can be written as
\begin{align}\label{eq:7}
    {W_F}
    = \mathop {\sup }\limits_{{{\left\| {{{\left\| C \right\|}_F}} \right\|}_L} \le K} {\expect_{{{\tilde {\cal D}}_s}}}\left[ {{{\left\| {C\left( f \right)} \right\|}_F}} \right] - {\expect_{{{\tilde {\cal D}}_t}}}\left[ {{{\left\| {C\left( f \right)} \right\|}_F}} \right],
\end{align}
where $W_F$ is short for ${W_F}\left( {{{\tilde {\cal D}}_s},{{\tilde {\cal D}}_t}} \right)$, which denotes the Frobenius norm-based 1-Wasserstein distance of two domain distributions. In this way, we can achieve explicit domain alignment and category distinguishment through a unified objective, contributing to leveraging the predicted discriminative information for capturing the multi-modal structures of the feature distributions.

\subsection{Adversarial Learning with the NWD}\label{sec:3.3}
\noindent\textbf{From Frobenius Norm to Nuclear Norm.} The constructed discriminator/critic $D = {\left\| C \right\|_F}$ can perform adversarial training with the generator $G$, which helps achieve transferable and discriminative representations while improving the prediction determinacy. However, adversarial learning based on the Frobenius-norm 1-Wasserstein distance may reduce the prediction diversity because it tends to push the category with a small number of samples to the neighbouring category containing large amounts of samples far from the decision boundary \cite{cui2021fast}. Inspired by recent works on the Nuclear norm \cite{recht2010guaranteed,cui2020towards,srebro2004maximum,cui2021fast}, which has been demonstrated to be bound with the Frobenius norm, we attempt to replace the Frobenius norm ${\left\|  \cdot  \right\|_F}$ with the Nuclear norm ${\left\|  \cdot  \right\|_*}$ because maximizing ${\left\| Z \right\|_*}$ means maximizing the rank of $Z$ when the ${\left\|  \cdot  \right\|_F}$ is nearby $\sqrt b$ \cite{cui2020towards,cui2021fast}, which  improves the prediction diversity. Therefore, the domain discrepancy can be rewritten as
\begin{align}\label{eq:8}
    {W_N}
    = \mathop {\sup }\limits_{{{\left\| {{{\left\| C \right\|}_*}} \right\|}_L} \le K} {\expect_{{{\tilde {\cal D}}_s}}}\left[ {{{\left\| {C\left( f \right)} \right\|}_*}} \right] - {\expect_{{{\tilde {\cal D}}_t}}}\left[ {{{\left\| {C\left( f \right)} \right\|}_*}} \right],
\end{align}
where $W_N$ is short for ${W_N}\left( {{{\tilde {\cal D}}_s},{{\tilde {\cal D}}_t}} \right)$, which denotes the Nuclear-norm 1-Wasserstein discrepancy (NWD) of two domain distributions. Then, our discriminator can be rewritten as $D = {\left\| C \right\|_*}$. When classifier $C$ is used for classification, it helps achieve category-level distinguishment, but when $C$ serves as a discriminator, it achieves feature-level alignment. Note that our classifier consists of a fully connected layer and a softmax activation function. It can be demonstrated that all the components of our implicit discriminator satisfy the K-Lipschitz constraint (\textbf{\textbf{see supplementary material for the proof}}), which enables the proposed model to be trained without the requirements of additional weight clipping and gradient penalty strategies. Therefore, we can approximately estimate the empirical NWD ${\hat W_N}$ by maximizing the domain critic loss ${\cal L}_{nwd}$:
\begin{equation}\label{eq:9}
        {{\cal L}}_{nwd}\left( {{x^s},{x^t}} \right) = \frac{1}{{{N_s}}}{\sum\limits_{i = 1}^{{N_s}} D\left(G\left( {x_i^s} \right)\right)} - \frac{1}{{{N_t}}}{\sum\limits_{j = 1}^{{N_t}}{D\left(G\left( {x_i^t} \right)\right)}},
\end{equation}
\begin{equation}\label{eq:10}
    {\hat W_N} = \mathop {\max }\limits_D {{\cal L}}_{nwd}\left( {{x^s},{x^t}} \right).
\end{equation}

\noindent\textbf{Adversarial Learning for DALN.} In this work, we build a DALN consisting of a generator $G$ based on a pretrained ResNet and a classifier $C$ constructed with a fully connected layer and a softmax layer. To avoid tedious alternating updates for the DALN, a gradient reverse layer (GRL) \cite{ganin2016domain}, which does not include the above mentioned gradient penalty or weight clipping, is used to help achieve updating within one back propagation. In this way, DALN can be trained by playing the min-max game as
\begin{equation}\label{eq:11}
    \mathop {\min }\limits_G \mathop {\max }\limits_C {{\cal L}}_{nwd}\left( {{x^s},{x^t}} \right).
\end{equation}
Moreover, to ensure the fidelity of UDA classification, we need to guarantee a low source risk for the source domain. Therefore, generator $G$ and classifier $C$ should also be  optimized by minimizing the supervised classification loss ${\cal L}_{cls}$ for the source domain as
\begin{equation}\label{eq:12}
    {\cal L}_{cls} \left( {{x^s},{y^s}} \right) = \frac{1}{{{N_s}}}\sum\limits_{i = 1}^{{n_s}} {{\cal L}_{ce}}\left( {C\left( {G\left( {x_i^s} \right)} \right),y_i^s} \right).
\end{equation}
In short, the overall loss used to optimize the classification model can be written as
\begin{equation}\label{eq:13}
    \mathop {\min }\limits_{C,G} \left\{ {{{\cal L}_{cls}}\left( {{x^s},{y^s}} \right) + \lambda \mathop {\max }\limits_C {{\cal L}}_{nwd}\left( {{x^s},{x^t}} \right)} \right\},
\end{equation}
where $\lambda$ is used to balance ${\cal L}_{cls}$ and ${\cal L}_{nwd}$. In this work, $\lambda$ is set to 1.  With the help of adversarial learning, the DALN learns transferable and discriminative representations while promising the prediction determinacy and diversity. 

\begin{table*}[htbp]\centering
\vspace{-1mm} 
    \caption{Classification accuracy (\%) on (a) Office-Home  and (b) VisDA-2017  for unsupervised domain adaptation (using ResNet-50 and ResNet-101 as the backbone, respectively). $^\dag$ denotes that the results are reproduced using the publicly released  code. The best accuracy is indicated in \textbf{\textcolor{red}{bold red}} and the second best accuracy is indicated in \textcolor{blue}{{\ul undelined blue}}. See \textbf{supplementary material} for more details.}\label{tab:home_visda}
    \vspace{-2mm}
	\captionsetup[subfloat]{}
	\captionsetup[subfloat]{justification=centering}
	\subfloat[Office-Home.\label{tab:home}]{
		\resizebox{0.75\textwidth}{!}{
            \begin{tabular}{lccccccccccccc}
                \hline
                Method&A→C&A→P&A→R&C→A&C→P&C→R&P→A&P→C&P→R&R→A&R→C&R→P&Avg\\
                \hline
                ResNet-50\cite{he2016deep}&34.9&50.0&58.0&37.4&41.9&46.2&38.5&31.2&60.4&53.9&41.2&59.9&46.1\\
                WDGRL$^\dag$(18)\cite{shen2018wasserstein}&44.1&63.8&74.0&47.3&57.1&61.7&51.8&39.1&72.1&64.9&45.9&76.5&58.2\\
                MCD(18)\cite{saito2018maximum}&48.9&68.3&74.6&61.3&67.6&68.8&57.0&47.1&75.1&69.1&52.2&79.6&64.1\\
                BSP(19)\cite{chen2019transferability}&52.0&68.6&76.1&58.0&70.3&70.2&58.6&50.2&77.6&72.2&59.3&81.9&66.3\\
                BNM(20)\cite{cui2020towards}&52.3&73.9&80.0&63.3&72.9&74.9&61.7&49.5&79.7&70.5&53.6&82.2&67.9\\
                GVB-GD(20)\cite{cui2020gradually}&57.0&74.7&79.8&64.6&74.1&74.6&65.2&55.1&81.0&{\color[HTML]{FE0000}\textbf{74.6}}&59.7&84.3&70.4\\
                FGDA(21)\cite{gao2021gradient}&52.3&77.0&78.2&64.6&75.5&73.7&64.0&49.5&80.7&70.1&52.3&81.6&68.3\\
                TSA(21)\cite{li2021transferable}&53.6&75.1&78.3&64.4&73.7&72.5&62.3&49.4&77.5&72.2&58.8&82.1&68.3\\
                CKB-MMD(21)\cite{luo2021conditional}&54.2&74.1&77.5&64.6&72.2&71.0&64.5&53.4&78.7&72.6&58.4&82.8&68.7\\
                SCDA(21)\cite{li2021semantic}&57.5&76.9&80.3&65.7&74.9&74.5&65.5&53.6&79.8&{\color[HTML]{0000FF}{\ul74.5}}&59.6&83.7&70.5\\
                MetaAlign(21)\cite{wei2021metaalign}&{\color[HTML]{FE0000}\textbf{59.3}}&76.0&80.2&65.7&74.7&75.1&65.7&{\color[HTML]{FE0000}\textbf{56.5}}&{\color[HTML]{0000FF}{\ul81.6}}&74.1&{\color[HTML]{0000FF}{\ul61.1}}&85.2&71.3\\
                \hline
                \textbf{DALN(Ours)}&57.8&{\color[HTML]{FE0000}\textbf{79.9}}&{\color[HTML]{0000FF}{\ul82.0}}&{\color[HTML]{0000FF}{\ul66.3}}&{\color[HTML]{0000FF}{\ul76.2}}&{\color[HTML]{0000FF}{\ul77.2}}&{\color[HTML]{0000FF}{\ul66.7}}&55.5&81.3&73.5&60.4&{\color[HTML]{0000FF}{\ul85.3}}&{\color[HTML]{0000FF}{\ul71.8}}\\
                \hline
                DANN(16)\cite{ganin2016domain}&45.6&59.3&70.1&47.0&58.5&60.9&46.1&43.7&68.5&63.2&51.8&76.8&57.6\\
                DANN+\textbf{NWD}&51.8&63.3&73.9&56.6&66.1&68.6&59.3&54.6&79.0&70.5&{\color[HTML]{FE0000}\textbf{61.5}}&80.4&65.5\\\hline
                CDAN(18)\cite{NEURIPS2018_ab88b157}&50.7&70.6&76.0&57.6&70.0&70.0&57.4&50.9&77.3&70.9&56.7&81.6&65.8\\
                CDAN+\textbf{NWD}&54.8&70.7&77.9&60.5&69.6&71.8&61.2&55.0&80.9&{\color[HTML]{FE0000}\textbf{74.6}}&59.4&83.4&68.3\\\hline
                MDD(19)\cite{zhang2019bridging}&54.9&73.7&77.8&60.0&71.4&71.8&61.2&53.6&78.1&72.5&60.2&82.3&68.1\\
                MDD+\textbf{NWD}&55.8&76.1&79.1&64.3&73.3&73.2&63.6&55.0&80.2&73.8&{\color[HTML]{0000FF}{\ul61.1}}&84.0&70.0\\
                \hline
                MCC(20)\cite{jin2020minimum}&55.1&75.2&79.5&63.3&73.2&75.8&66.1&52.1&76.9&73.8&58.4&83.6&69.4\\
                MCC+\textbf{NWD}&{\color[HTML]{0000FF}{\ul58.1}}&{\color[HTML]{0000FF}{\ul79.6}}&{\color[HTML]{FE0000}\textbf{83.7}}&{\color[HTML]{FE0000}\textbf{67.7}}&{\color[HTML]{FE0000}\textbf{77.9}}&{\color[HTML]{FE0000}\textbf{78.7}}&{\color[HTML]{FE0000}\textbf{66.8}}&{\color[HTML]{0000FF}{\ul56.0}}&{\color[HTML]{FE0000}\textbf{81.9}}&73.9&60.9&{\color[HTML]{FE0000}\textbf{86.1}}&{\color[HTML]{FE0000}\textbf{72.6}}\\
                \hline
            \end{tabular}
        }
    }
    \hspace{0.5mm}
	\subfloat[VisDA-2017.\label{tab:visda}]{
		\resizebox{0.217\textwidth}{!}{
    	    \begin{tabular}{lc}
                \hline
                Method&Avg\\
                \hline
                ResNet-101\cite{he2016deep}&52.4\\
                WDGRL$^\dag$(18)\cite{shen2018wasserstein}&61.3\\
                MCD(18)\cite{saito2018maximum}&71.9\\
                BSP(19)\cite{chen2019transferability}&75.9\\
                SWD(19)\cite{lee2019sliced}&76.4\\
                BNM(20)\cite{cui2020towards}&70.4\\
                GVB-GD$^\dag$(20)\cite{cui2020gradually}&77.2\\
                DADA(20)\cite{tang2020discriminative}&79.8\\
                TSA(21)\cite{li2021transferable}&78.6\\
                SCDA$^\dag$(21)\cite{li2021semantic}&79.7\\
                \hline
                \textbf{DALN(Ours)}&80.6\\
                \hline
                DANN(16)\cite{ganin2016domain}&57.4\\
                DANN+\textbf{NWD}&80.0(22.6↑)\\
                \hline
                CDAN(18)\cite{NEURIPS2018_ab88b157}&73.9\\
                CDAN+\textbf{NWD}&81.4(7.5↑)\\
                \hline
                MDD$^\dag$(19)\cite{zhang2019bridging}&76.8\\
                MDD+\textbf{NWD}&{\color[HTML]{0000FF} \ul{82.0(5.2↑)}}\\
                \hline
                MCC(20)\cite{jin2020minimum}&78.8\\
                MCC+\textbf{NWD}&{\color[HTML]{FE0000}\textbf{83.7(4.9↑)}}\\
                \hline
            \end{tabular}
        }
	}
	\hspace{0.5mm}
\end{table*}
\begin{table*}[htbp]\centering
\vspace{-1mm}
    \caption{Classification accuracy (\%) on (a) Office-31 and (b) ImageCLEF-2014 for unsupervised domain adaptation (using ResNet-50 as the backbone). $^\dag$ denotes that the results are reproduced using the publicly released  code. The best accuracy is indicated in \textbf{\textcolor{red}{bold red}} and the second best is indicated in \textcolor{blue}{{\ul undelined blue}}.\label{tab:31image}}
    \vspace{-2mm}
	\captionsetup[subfloat]{}
	\captionsetup[subffloat]{justification=centering}
	\subfloat[Office-31.\label{tab:31}]{
		\resizebox{0.487\textwidth}{!}{
            \begin{tabular}{lccccccc}
                \hline
                Method&A→W&D→W&W→D&A→D&D→A&W→A&Avg\\
                \hline
                ResNet-50\cite{he2016deep}&68.4&96.7&99.3&68.9&62.5&60.7&76.1\\
                WDGRL$^\dag$(18)\cite{shen2018wasserstein}&72.6&97.1&99.2&79.5&63.7&59.5&78.6\\
                MCD(18)\cite{saito2018maximum}&88.6&98.5&{\color[HTML]{FE0000}\textbf{100.0}}&92.2&69.5&69.7&86.5\\
                SWD(19)\cite{lee2019sliced}&90.4&98.7&{\color[HTML]{FE0000}\textbf{100.0}}&94.7&70.3&70.5&87.4\\
                BNM(20)\cite{cui2020towards}&91.5&98.5&{\color[HTML]{FE0000}\textbf{100.0}}&90.3&70.9&71.6&87.1\\
                DADA(20)\cite{tang2020discriminative}&92.3&{\color[HTML]{FE0000}\textbf{99.2}}&{\color[HTML]{FE0000}\textbf{100.0}}&93.9&74.4&74.2&89.0\\
                GVB-GD(20)\cite{cui2020gradually}&94.8&98.7&{\color[HTML]{FE0000}\textbf{100.0}}&95.0&73.4&73.7&89.3\\
                FGDA(21)\cite{gao2021gradient}&93.3&99.1&{\color[HTML]{FE0000}\textbf{100.0}}&93.2&73.2&72.7&88.6\\
                MetaAlign(21)\cite{wei2021metaalign}&93.0&98.6&{\color[HTML]{FE0000}\textbf{100.0}}&94.5&75.0&73.6&89.2\\
                TSA(21)\cite{li2021transferable}&94.8&99.1&{\color[HTML]{FE0000}\textbf{100.0}}&92.6&74.9&74.4&89.3\\
                SCDA(21)\cite{li2021semantic}&94.2&98.7&{\color[HTML]{333333}99.8}&{\color[HTML]{0000FF}{\ul95.2}}&75.7&{\color[HTML]{0000FF}{\ul76.2}}&{\color[HTML]{0000FF}{\ul90.0}}\\
                \hline
                \textbf{DALN(Ours)}&{\color[HTML]{0000FF}{\ul95.2}}&{\color[HTML]{0000FF}{\ul99.1}}&{\color[HTML]{FE0000}\textbf{100.0}}&{\color[HTML]{FE0000}\textbf{95.4}}&{\color[HTML]{0000FF}{\ul76.4}}&{\color[HTML]{FE0000}\textbf{76.5}}&{\color[HTML]{FE0000}\textbf{90.4}}\\
                \hline
                DANN(16)\cite{ganin2016domain}&82.0&96.9&99.1&79.7&68.2&67.4&82.2\\
                DANN+\textbf{NWD}&92.1&98.2&{\color[HTML]{FE0000}\textbf{100.0}}&84.7&74.5&73.0&87.1\\
                \hline
                CDAN(18)\cite{NEURIPS2018_ab88b157}&94.1&98.6&{\color[HTML]{FE0000}\textbf{100.0}}&92.9&71.0&69.3&87.7\\
                CDAN+\textbf{NWD}&93.7&98.5&{\color[HTML]{FE0000}\textbf{100.0}}&91.0&74.4&73.0&88.4\\
                \hline
                MDD(19)\cite{zhang2019bridging}&94.5&98.4&{\color[HTML]{FE0000}\textbf{100.0}}&93.5&74.6&72.2&88.9\\
                MDD+\textbf{NWD}&{\color[HTML]{FE0000}\textbf{95.5}}&98.7&{\color[HTML]{FE0000}\textbf{100.0}}&94.9&{\color[HTML]{FE0000}\textbf{76.6}}&74.0&{\color[HTML]{0000FF}{\ul90.0}}\\
                \hline
                MCC(20)\cite{jin2020minimum}&{\color[HTML]{FE0000}\textbf{95.5}}&98.6&{\color[HTML]{FE0000}\textbf{100.0}}&94.4&72.9&74.9&89.4\\
                MCC+\textbf{NWD}&{\color[HTML]{FE0000}\textbf{95.5}}&98.7&{\color[HTML]{FE0000}\textbf{100.0}}&{\color[HTML]{FE0000}\textbf{95.4}}&75.0&75.1&{\color[HTML]{0000FF}{\ul90.0}}\\
                \hline
            \end{tabular}
        }
    }
    \hspace{0.5mm}
	\subfloat[ImageCLEF-2014. \label{tab:image}]{
		\resizebox{0.469\textwidth}{!}{
    	    \begin{tabular}{lccccccc}
                \hline
                Method&I→P&P→I&I→C&C→I&C→P&P→C&Avg\\
                \hline
                ResNet-50\cite{he2016deep}&74.8&83.9&91.5&78.0&65.5&91.2&80.7\\
                WDGRL$^\dag$(18)\cite{shen2018wasserstein}&76.8&87.0&91.7&87.2&75.2&90.3&84.7\\
                MCD(18)\cite{saito2018maximum}&77.3&89.2&92.7&88.2&71.0&92.3&85.1\\
                SWD(19)\cite{lee2019sliced}&78.1&89.6&95.2&89.3&73.4&92.8&86.4\\
                BNM(20)\cite{cui2020towards}&77.2&91.2&96.2&91.7&75.7&{\color[HTML]{0000FF}{\ul96.7}}&88.1\\
                GVB-GD$^\dag$(20)\cite{cui2020gradually}&78.2&92.7&96.5&91.5&78.2&95.0&88.7\\
                DADA$^\dag$(20)\cite{tang2020discriminative}&78.7&92.3&97.2&91.6&78.5&95.3&88.9\\
                CKB-MMD(21)\cite{luo2021conditional}&{\color[HTML]{FE0000}\textbf{80.7}}&92.2&96.5&92.2&{\color[HTML]{0000FF}{\ul79.9}}&{\color[HTML]{0000FF}{\ul96.7}}&{\color[HTML]{0000FF}{\ul89.7}}\\
                SCDA$^\dag$(21)\cite{li2021semantic}&78.7&91.8&96.7&92.8&78.5&95.2&89.0\\
                TSA$^\dag$(21)\cite{li2021transferable}&78.6&92.8&97.0&92.8&79.0&95.2&89.2\\
                \hline
                \textbf{DALN(Ours)}&{\color[HTML]{0000FF}{\ul80.5}}&{\color[HTML]{0000FF}{\ul93.8}}&{\color[HTML]{0000FF}{\ul97.5}}&92.8&78.3&95.0&{\color[HTML]{0000FF}{\ul89.7}}\\
                \hline
                DANN(16)\cite{ganin2016domain}&75.0&86.0&96.2&87.0&74.3&91.5&85.0\\
                DANN+\textbf{NWD}&78.0&89.2&97.3&{\color[HTML]{0000FF}{\ul93.3}}&78.5&92.0&88.1\\
                \hline
                CDAN(18)\cite{NEURIPS2018_ab88b157}&77.7&90.7&97.7&91.3&74.2&94.3&87.7\\
                CDAN+\textbf{NWD}&78.6&92.5&97.2&91.7&79.3&94.6&89.0\\
                \hline
                MDD$^\dag$(19)\cite{zhang2019bridging}&77.3&90.2&96.8&89.5&77.6&94.2&87.6\\
                MDD+\textbf{NWD}&78.9&91.7&{\color[HTML]{0000FF}{\ul97.5}}&91.7&78.9&95.4&89.0\\
                \hline
                MCC$^\dag$(20)\cite{jin2020minimum}&78.3&{\color[HTML]{FE0000}\textbf{94.5}}&97.3&92.3&77.3&96.3&89.3\\
                MCC+\textbf{NWD}&79.8&{\color[HTML]{FE0000}\textbf{94.5}}&{\color[HTML]{FE0000}\textbf{98.0}}&{\color[HTML]{FE0000}\textbf{94.2}}&{\color[HTML]{FE0000}\textbf{80.0}}&{\color[HTML]{FE0000}\textbf{97.5}}&{\color[HTML]{FE0000}\textbf{90.7}}\\
                \hline
            \end{tabular}
        }
    }
    \hspace{0.5mm}
\vspace{-2mm}
\end{table*}

\noindent\textbf{Generalization Bound.} Here, we present the theoretical guarantees for the proposed method. Following \cite{ben2007analysis}, we consider a binary classification instance. Then, let $\cal F$ $\left(f \in \cal F \right)$ denote a fixed representation space and $C:{\cal F} \to \left[ {0,1} \right]$ be a family of source classifiers, where $C$ belongs to hypothesis space $\cal H$. We assume that the risk of $C$ on the source domain is described as ${\varepsilon _s}\left( C \right) = {\expect_{f \sim {{\tilde {\cal D}}_s}}}\left[ {C\left( f \right) \ne y} \right]$, where $\caldfs{}$ is the feature distribution induced by the data distribution of source domain $\calds$ and $y$ is the label corresponding to the induced feature $f$. Moreover, given two classifiers ${C_1},{C_2} \in {\cal H}$, we define the risk of these two classifiers on the source domain as ${\varepsilon _s}\left( {{C_1},{C_2}} \right) = {\expect_{f \sim {{\tilde {\cal D}}_s}}}\left[ {{C_1}\left( f \right) \ne {C_2}\left( f \right)} \right]$. In the same way, we define the risk on the target domain, i.e, ${\varepsilon _t}\left( C \right)$ and ${\varepsilon _t}\left( {{C_1},{C_2}} \right)$. Then, the ideal joint hypothesis is written as ${C^ * } = \arg \mathop {\min }\limits_C {\varepsilon _s}\left( C \right) + {\varepsilon _t}\left( C \right)$, which can be used to minimize the combined risk on the source and target domains. Therefore, according to \cite{ben2007analysis}, the probabilistic bound of ${\varepsilon _t}\left( C \right)$ can be written as
\begin{equation}\label{eq:14}
    {\varepsilon _t}\left( C \right) \le {\varepsilon _s}\left( C \right) + \left| {{\varepsilon _s}\left( {C,{C^ * }} \right) - {\varepsilon _t}\left( {C,{C^ * }} \right)} \right| + \eta^*,
\end{equation}
where $\eta^*={{\varepsilon _s}\left( {{C^ * }} \right) + {\varepsilon _t}\left( {{C^ * }} \right)}$ is a sufficiently small constant representing the ideal combined risk. Thus, the goal of UDA classification is to reduce the domain discrepancy term $\left| {{\varepsilon _s}\left( {C,{C^ * }} \right) - {\varepsilon _t}\left( {C,{C^ * }} \right)} \right|$.
\begin{lemma} \label{lemma1}
Let ${\nu_s},{\nu_t} \in \cal P \left( {\cal F} \right)$ denote the probability measures of the source and target domain features, $\rho \left( {{f^s},{f^t}} \right)$ be the cost of transporting a unit of material from location $f^s$ satisfying $f^s \sim {\nu_s}$ to location $f^t$ satisfying $f^t \sim {\nu_t}$, ${W_1}\left( {{\nu_s},{\nu_t}} \right)$ represent the NWD, and $K$ denote a Lipschitz constant. Given a family of classifiers $C \in {\cal H}_1$ and a ideal classifier ${C^ * } \in {{\cal H}_1}$ satisfying the K-Lipschitz constraint, where ${\cal H}_1$ is a subspace of $\cal H$, the following holds for every $C,{C^ * } \in {{\cal H}_1}$.
\begin{equation}\label{eq:15}
    \left| {{\varepsilon _s}\left( {C,{C^ * }} \right) - {\varepsilon _t}\left( {C,{C^ * }} \right)} \right| \leqslant 2K{W_1}\left( {{\nu_s},{\nu_t}} \right).
\end{equation}
\end{lemma}
\begin{thm}\label{thm1}
Based on Lemma \ref{lemma1}, for every $C \in {{\cal H}_1}$, the following holds
\begin{equation}\label{eq:16}
    {\varepsilon _t}\left( C \right) \le {\varepsilon _s}\left( C \right) + 2K{W_1}\left( {{\nu_s},{\nu_t}} \right) + \eta^*, 
\end{equation}
where $\eta^*={{\varepsilon _s}\left( {{C^ * }} \right) + {\varepsilon _t}\left( {{C^ * }} \right)}$ is the risk of ideal joint hypothesis, which is a sufficiently small constant.
\end{thm}
Therefore, the risk of the target domain can be bounded by the risk of the source domain and the introduced NWD, providing theoretical guarantees for the proposed approach. Limited by space, \textbf{all the proofs and more details about the empirical measure of the target risk are provided in the supplementary materials.}

\subsection{Regularizer to Existing UDA Methods}\label{sec:3.4}
The proposed NWD can be easily integrated into existing methods to improve the prediction determinacy and diversity. Specifically, a gradient reverse layer is first added to the original task-specific classifier. Subsequently, the task-specific classifier with the introduced NWD can serve as a discriminator, which performs adversarial learning with the feature extractor. Formally, assuming the original loss ${\cal L}_{ori}$ of the model written as ${\cal L}_{ori} = {{\cal L}_{cls}}{\rm{ + }}{{\cal L}}_{spe}$ , where ${\cal L}_{cls}$ is the standard supervised classification loss as that of the proposed method and ${\cal L}_{spe}$ is the special loss used in these methods, the reconstructed loss ${\cal L}_{rec}$ can be described as
\begin{equation}\label{eq:17}
    {{\cal L}_{rec}} = {{\cal L}_{cls}}{\rm{ + }}{{\cal L}_{spe}} + \gamma {{\cal L}_{nwd}},
\end{equation}
where $\gamma$ is the balance weight. For convenience, in our experiments, the values of $\gamma$ for all other methods are set to 0.01. The results of taking the proposed discrepancy as a regularizer to benefit other UDA algorithms are presented in the experiments.

\section{Experiments}
In this section, we evaluate the proposed DALN and compare it with the SOTA methods for UDA classification. Additionally, we evaluate the effectiveness of NWD as a regularizer to benefit existing methods including DANN \cite{ganin2016domain}, CDAN \cite{NEURIPS2018_ab88b157}, MDD \cite{zhang2019bridging}, and MCC \cite{jin2020minimum}.

\subsection{Setup}
\noindent \textbf{Dataset Description}. We use four datasets including Office-Home \cite{venkateswara2017deep}, Office-31 \cite{saenko2010adapting}, ImageCLEF \cite{caputo2014imageclef}, and VisDA-2017 \cite{peng2017visda} to perform the comparison experiments. \textbf{Office-Home} is a large-scale dataset that includes 15500 images 
and 65 categories. This dataset has four extremely different domains, i.e., Art (A), Clip Art (C), Product (P), and Real-World (R). \textbf{VisDA-2017}  is a large-scale synthetic-to-real dataset that has two domains (synthetic (S) and real (R)). The dataset contains over 280K images across 12 classes. \textbf{Office-31} has a total number of 4110 images, which includes three domains, i.e., Amazon (A), Webcam (W), and DSLR (D); and each domain contains 31 categories. \textbf{ImageCLEF} consists of three domains derived from three public datasets: Caltech (C), ImageNet (I), and Pascal (P). Each domain contains 12 categories and each category has 50 images. 

\noindent\textbf{Implementation Details}. The proposed method is implemented based on the PyTorch \cite{paszke2019pytorch} framework running on a GPU (Tesla-V100). The SGD optimizer is used to train the model with a moment of 0.9, a weight decay of 1e-3, a batch size of 36, and a cropped image size of 224 × 224. The initial learning rate of classifier $C$ is set to 5e-3, which is 10 times larger than that of feature extractor $G$. \textbf{Other details can be found in the supplementary material.}

\subsection{Comparison Results}
\noindent\textbf{Results on Office-Home} are shown in \Table{home_visda}(a). Compared with SOTA methods, the proposed method achieves dramatic improvements in terms of classification accuracy. Particularly, in the case of domains suffering from large shifts and extremely unbalanced classes, e.g., A→R and C→R, the proposed method achieves 2.9\% and 2.2\% improvements compared to the existing SOTA methods. Moreover, by integrating the proposed NWD into the DANN, the average accuracy is improved by 7.9\%. Combining the proposed NWD with MCC, it achieves SOAT performance of 72.6\%, attaining 3.2\% improvements. Dramatic improvements are obviously exhibited in P→R, C→A, and C→P tasks. These obtained gains come from the paradigm leveraging the predicted discriminative information and the introduced NWD encouraging the prediction determinacy and diversity.

\noindent\textbf{Results on VisDA-2017} are displayed in \Table{home_visda}(b). Despite the tremendous domain shift existing in synthetic and real data, the DALN achieves an average accuracy of 80.6\%, outperforming the existing SOTA methods. Combining the proposed NWD with other methods, the performances of these methods are substantially improved by 22.6\%, 7.5\%, 5.2\%, and 4.9\% for the DANN, CDAN, MDD, and MCC, respectively. In particular, with the help of the proposed NWD, MCC achieves SOTA accuracy of 83.7\%, demonstrating the effectiveness of the proposed method.

\noindent\textbf{Results on Office-31} are presented in \Table{31image}(a). The proposed DALN achieves SOTA performances in five adaptation sub-tasks, and attains the best performance on the average accuracy. Particularly, compared with WDGRL \cite{shen2018wasserstein}, which uses 1-Wasserstein distance with an additional discriminator, the proposed DALN dramatically improves the average accuracy by 11.8\%. Additionally, taking the NWD as a regularizer, the typical methods can be improved by at least 0.6\%, and the average accuracy of DANN is even improved by 4.9\%. We note that the improvements are evidently achieved in the task of adapting a domain with a small number of samples (e.g., D and W) to a domain (e.g., A) containing large amounts of samples. These results occur because the proposed method encourages the prediction determinacy and diversity, which are highly important in such cases.

\noindent\textbf{Results on ImageCLEF-2014} are provided in \Table{31image}(b). Without bells and whistles, the proposed method achieves an average accuracy of 89.7\%, which is same as the most competitive method CKB-MMD \cite{luo2021conditional}. Moreover, the proposed method can be used as a regularizer, which is capable of improving the performance of the existing methods and thus contributes to SOTA performances. Specifically, the proposed method enables MCC to compare favorably against the SOTA methods. These results demonstrate that the proposed method is still highly effective in the case of all the domains containing the same samples and categories. 

\subsection{Insight Analysis}
 Here, we provide insight analyses for the proposed DALN and NWD. Limited by space, more detailed analyses regarding toy experiment, Proxy $\cal A$ distance, self-correlation matrix, convergence, and trade-off parameters ($\lambda$ and $\gamma$) are provided in \textbf{supplementary material.}
 
\begin{figure*}[htbp]
\centering
\subfloat[Source only]{\includegraphics[width=0.166\textwidth]{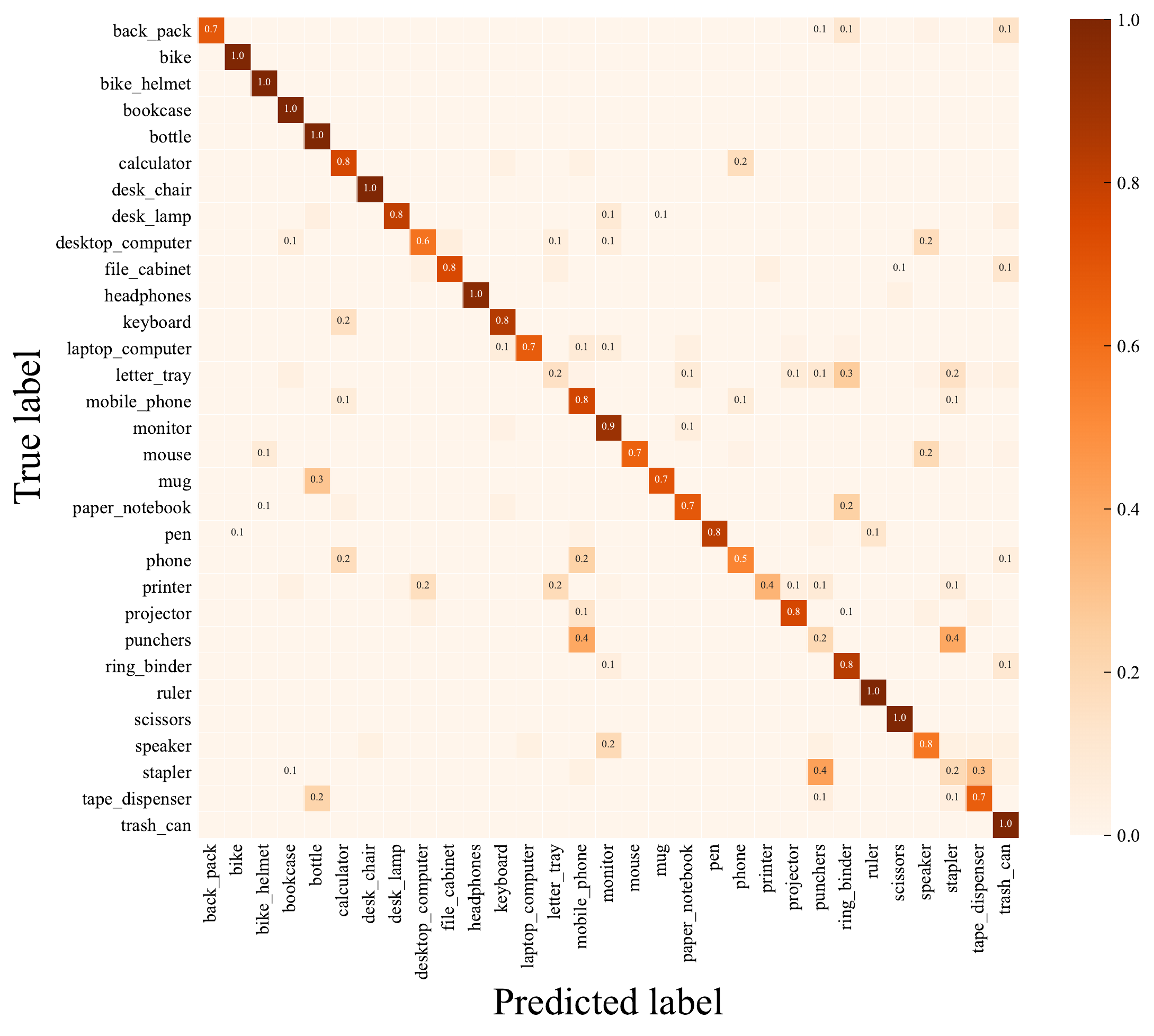}}
\subfloat[DANN]{\includegraphics[width=0.166\textwidth]{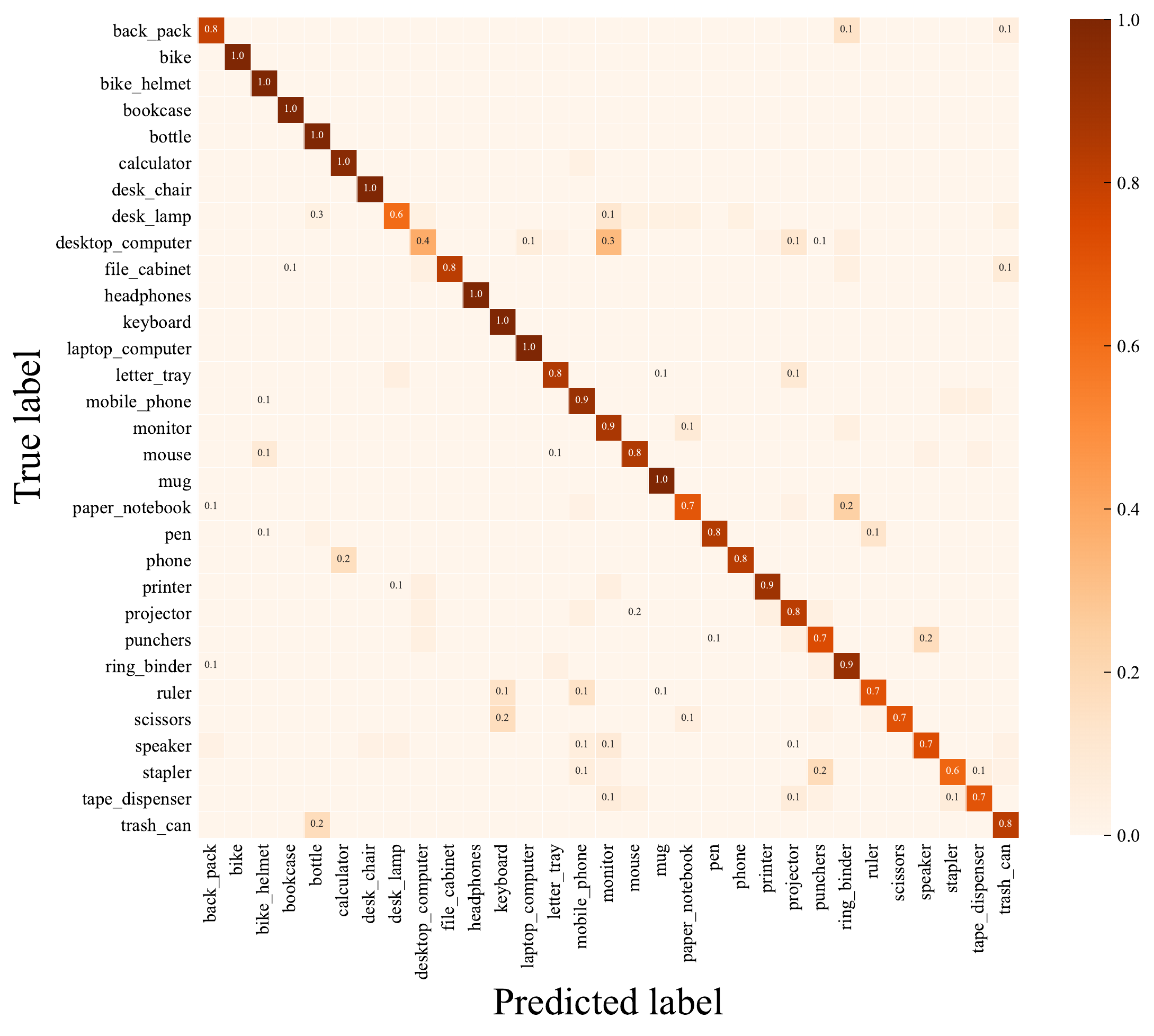}}
\subfloat[MDD]{\includegraphics[width=0.166\textwidth]{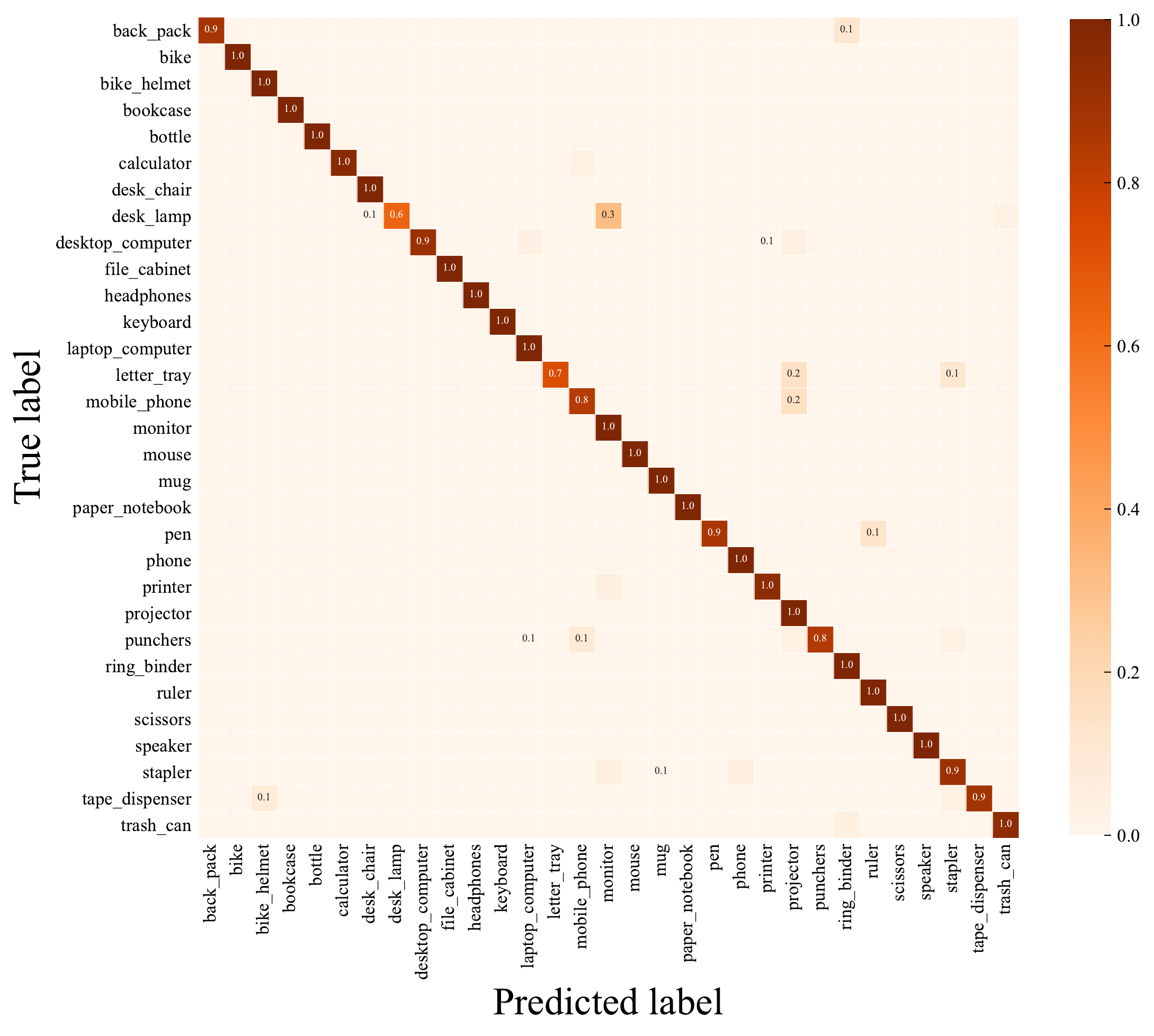}}
\subfloat[DALN]{\includegraphics[width=0.166\textwidth]{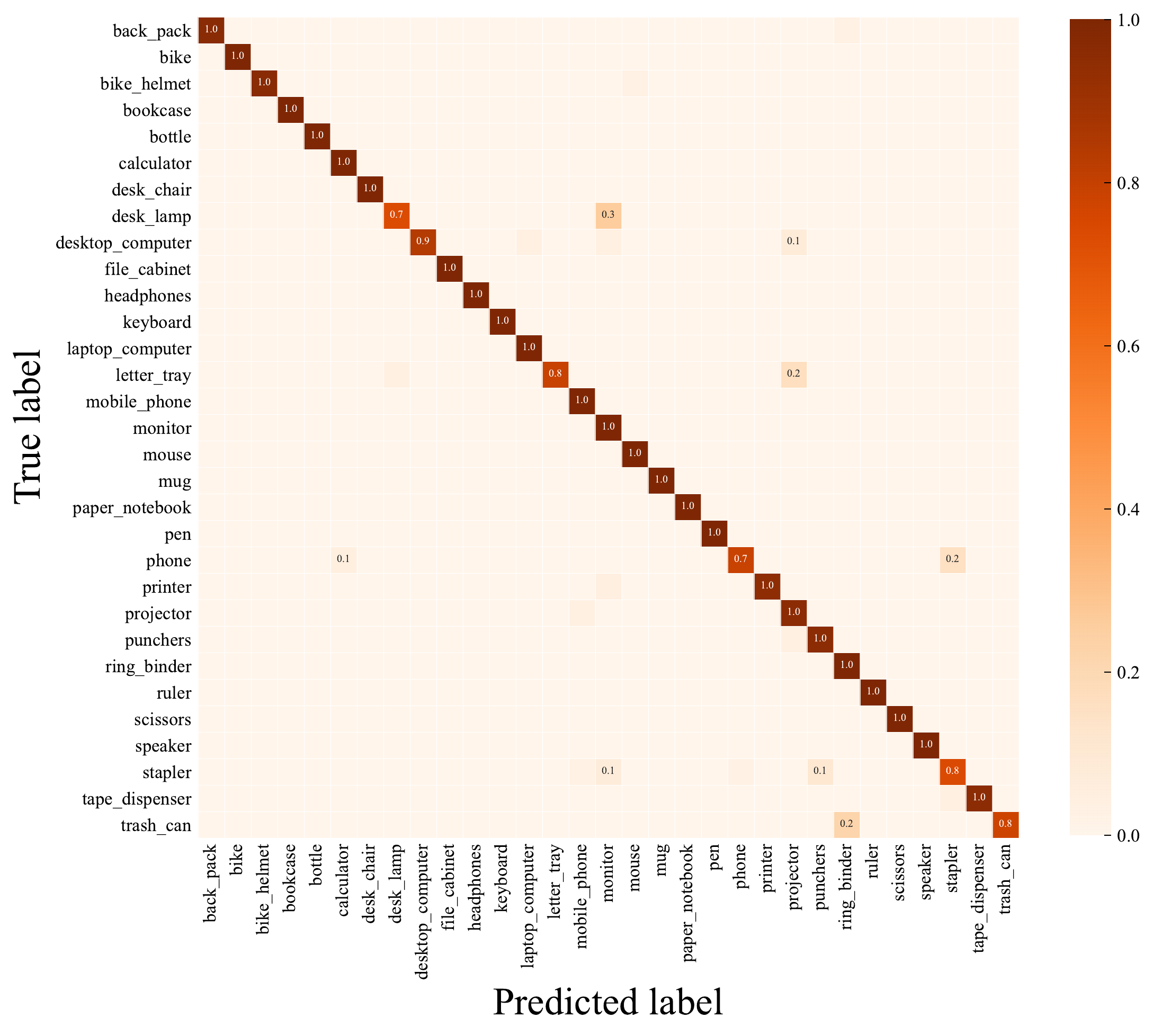}}
\subfloat[DANN+NWD]{\includegraphics[width=0.166\textwidth]{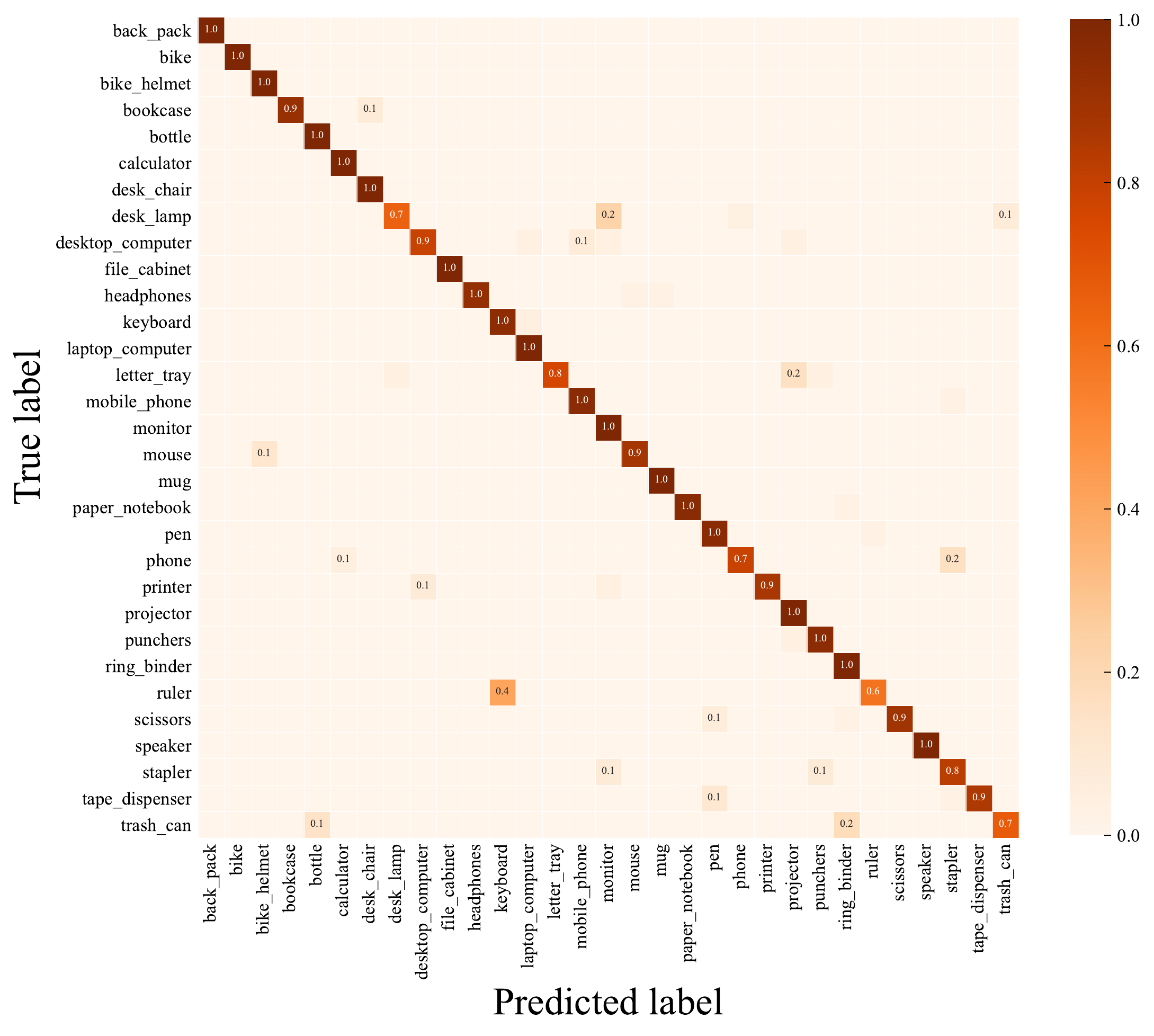}}
\subfloat[MDD+NWD]{\includegraphics[width=0.166\textwidth]{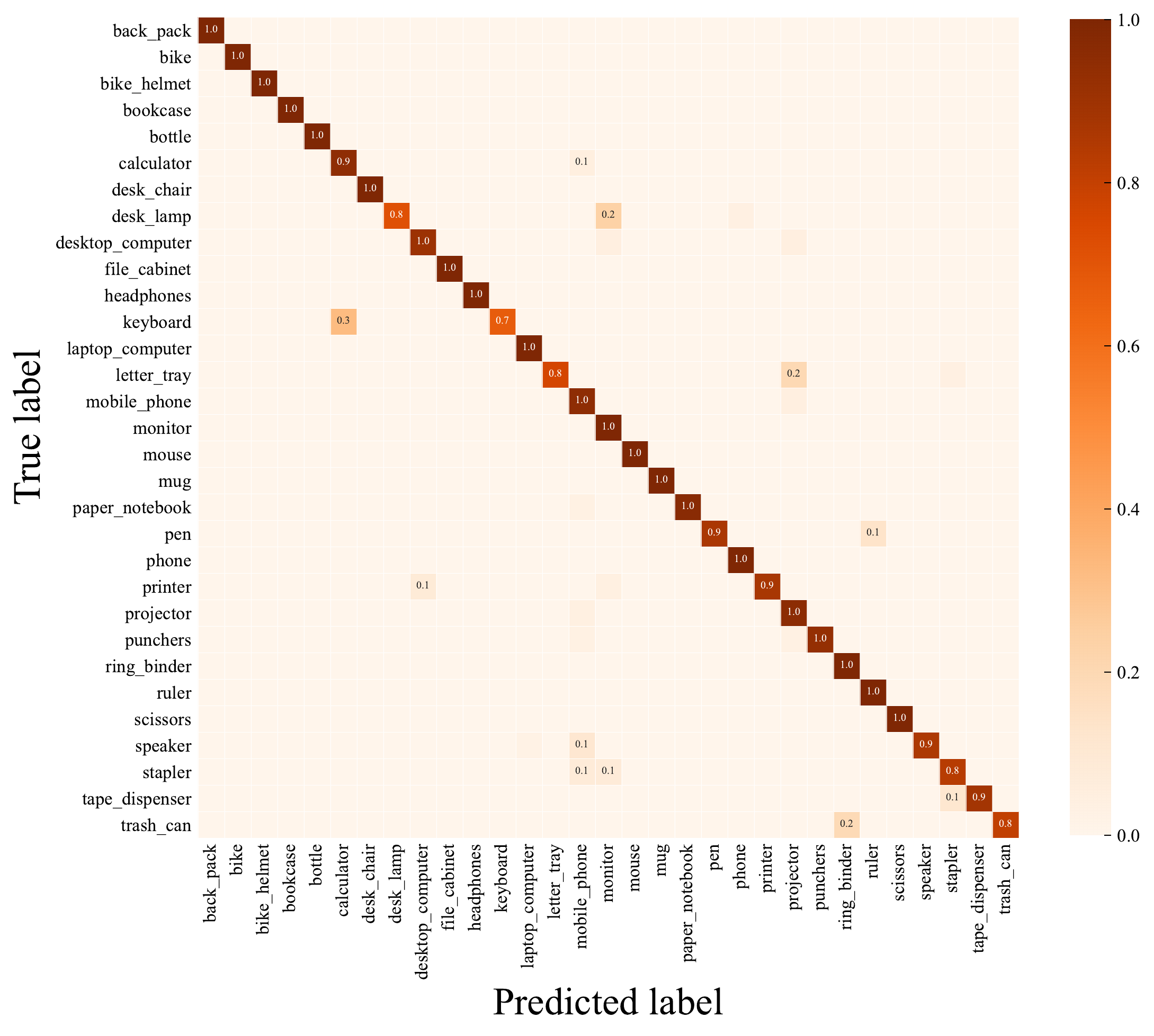}}
\vspace{-1mm}
\caption{
The confusion matrices of different methods of the target domain on task A → W of Office-31. (Zoom in for a clear visualization.)
}
\vspace{-4mm}
\label{fig:matrix}
\end{figure*}
\noindent\textbf{Confusion Matrix.} The comparison of the confusion matrices is shown in \Fig{matrix}. The figure shows that the model trained on the source-only data suffers from severe class confusion. The DANN focuses on domain-level feature adaptation, but ignores feature discriminability, which results in misclassification for some categories (e.g., computers are misclassified as monitors and projectors). In contrast, benefiting from the introduced paradigm, DALN generates large values for the main diagonal elements of the confusion matrix. Moreover, by integrating the NWD into DANN and MDD, the off-diagonal elements of their confusion matrix are considerably decreased, demonstrating the effectiveness of NWD.

\begin{figure}[htbp]
\centering
\vspace{-3mm}
\subfloat[Determinacy]{
    \includegraphics[width=0.48\linewidth]{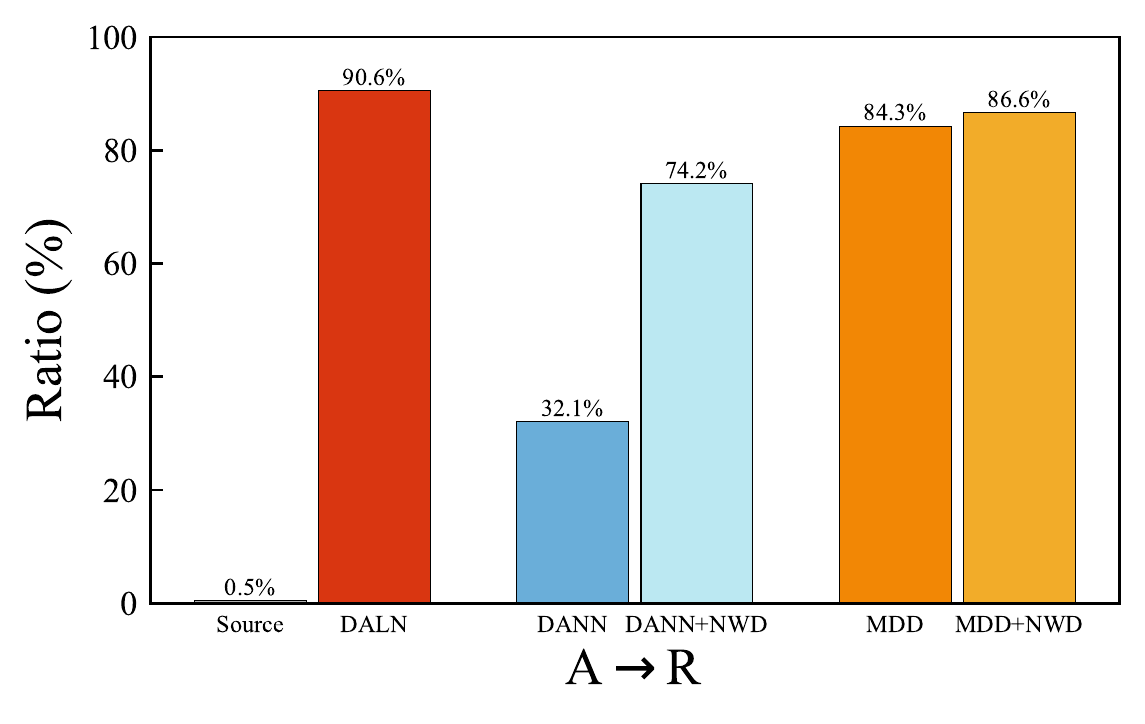}\hfill
}
\subfloat[Diversity]{
    {\includegraphics[width=0.48\linewidth]{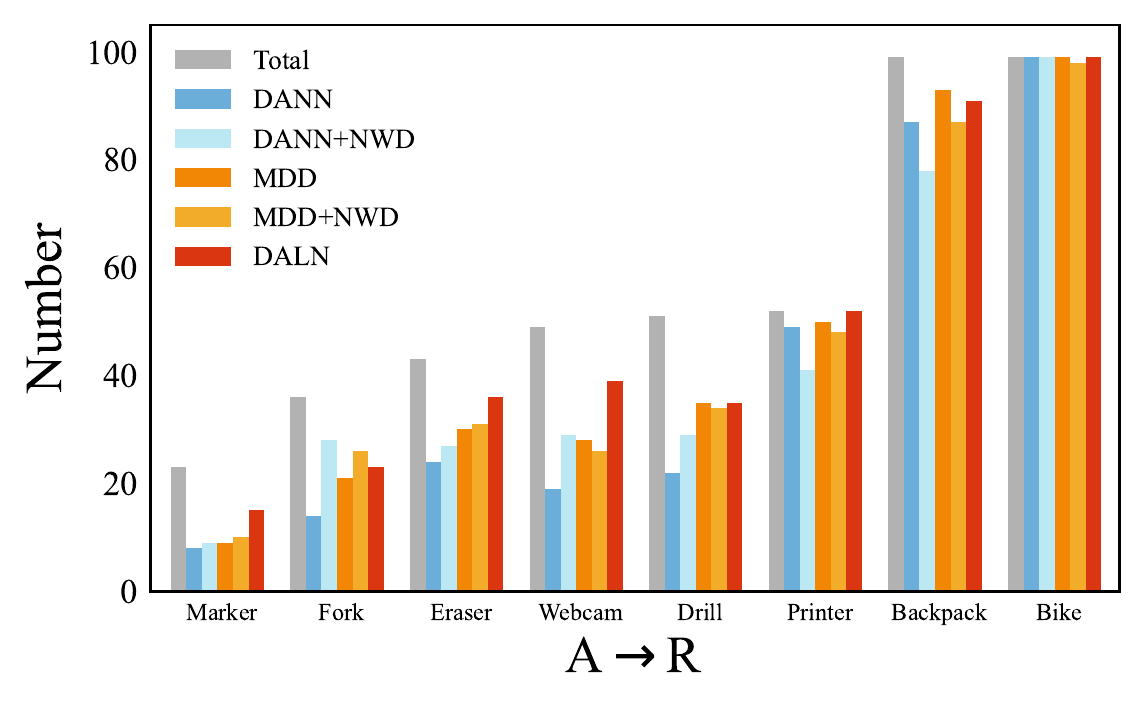}}\hfill
}
\vspace{-1mm}
\caption{
Visualizations of (a) determinacy and (b) diversity on task A→R of Office-Home. The ratio in (a) is the proportion of the number of correctly classified samples, whose prediction probability is in the range of 0.9 to 1, to the total number of correctly classified samples in the target domain. The number of correctly classified samples in (b) is calculated for 8 typical categories that have a large or small number of samples. 
}
\label{fig:bar}
\end{figure}

\noindent\textbf{Determinacy.} To evaluate the determinacy, we calculate the ratio of the correctly classified samples that have high prediction certainty. Here, we consider task Ar→Rw of Office-Home. The prediction probability in the range of 0.9 to 1 is regarded as a high certainty prediction. As shown in \Fig{bar}(a), the model trained on the source-only data nearly cannot generate high certainty prediction. DANN and MDD considerably improve the ratio of high certainty prediction, but the improvements achieved by these methods cannot compete with those achieved by the proposed DALN. By taking the NWD as a regularizer, the ratios of high certainty prediction for DANN and MDD are improved, demonstrating the effectiveness of NWD in improving the determinacy.

\noindent\textbf{Diversity.} As shown in \Fig{bar}(b), we compute the number of correctly classified samples for some typical categories that have a large or small number of samples. Compared with other methods, the proposed DALN correctly classifies more samples in the categories that have a small number of samples. Moreover, by adopting the NWD as a regularizer for the DANN and MDD, these methods achieve considerable improvements for categories that have a small number of samples. These results demonstrate the effectiveness of the NWD in improving prediction diversity.

\begin{figure}[htbp]
\centering
\vspace{-2mm}
\subfloat[Source only]{\includegraphics[width=0.32\linewidth,height=0.08\textheight]{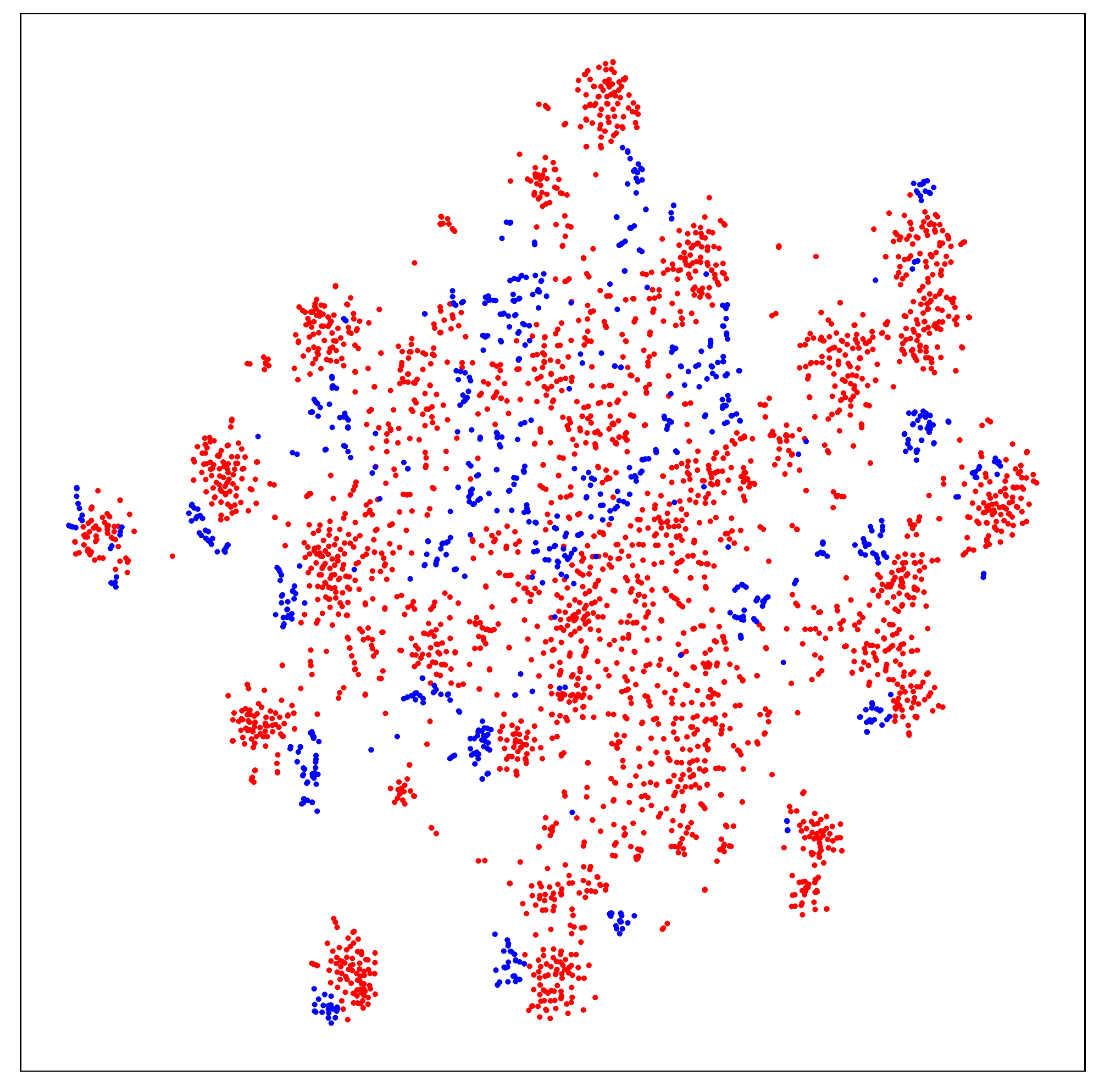}}
\subfloat[DANN]{\includegraphics[width=0.32\linewidth,height=0.08\textheight]{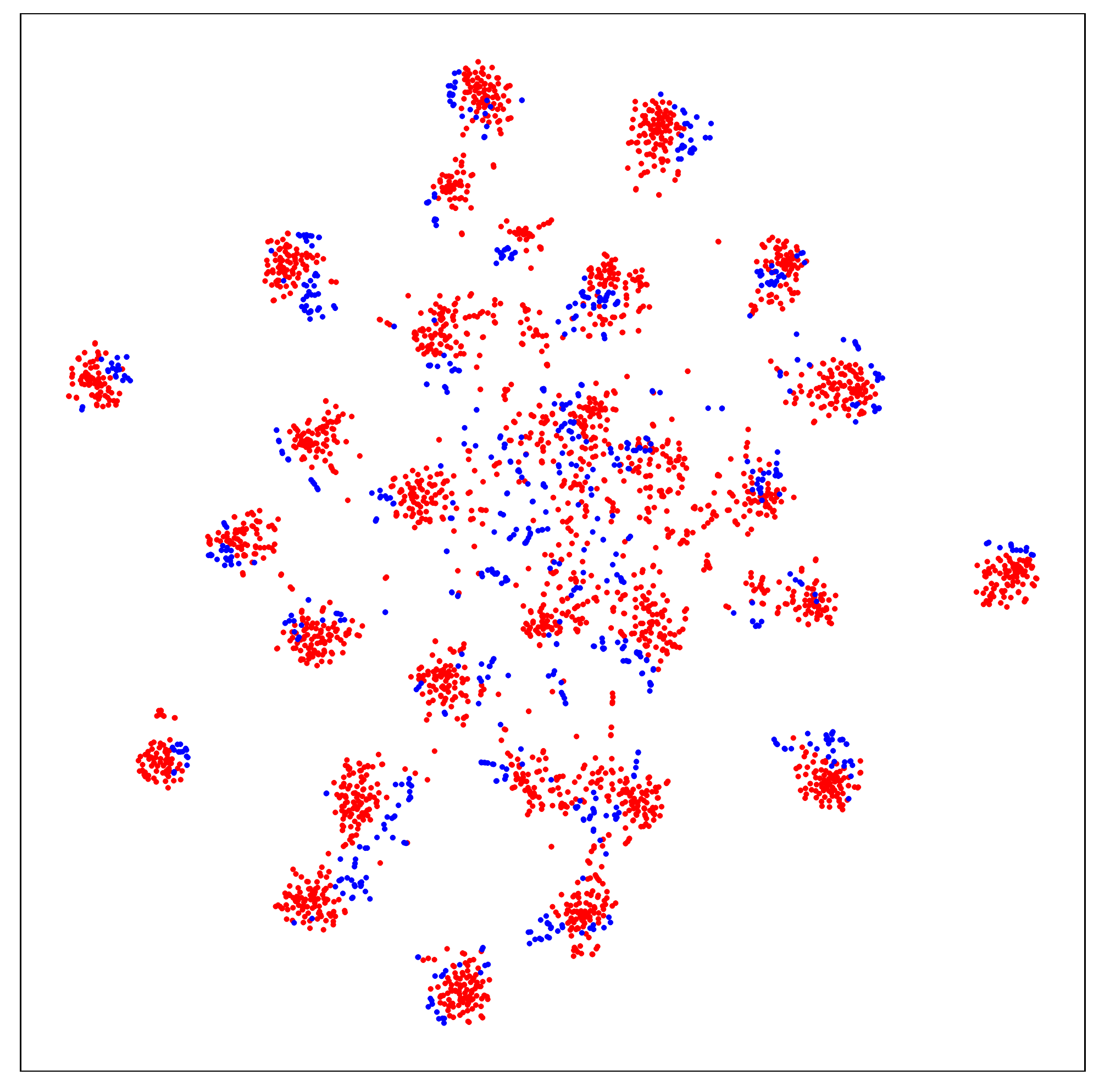}}
\subfloat[MDD]{\includegraphics[width=0.32\linewidth,height=0.08\textheight]{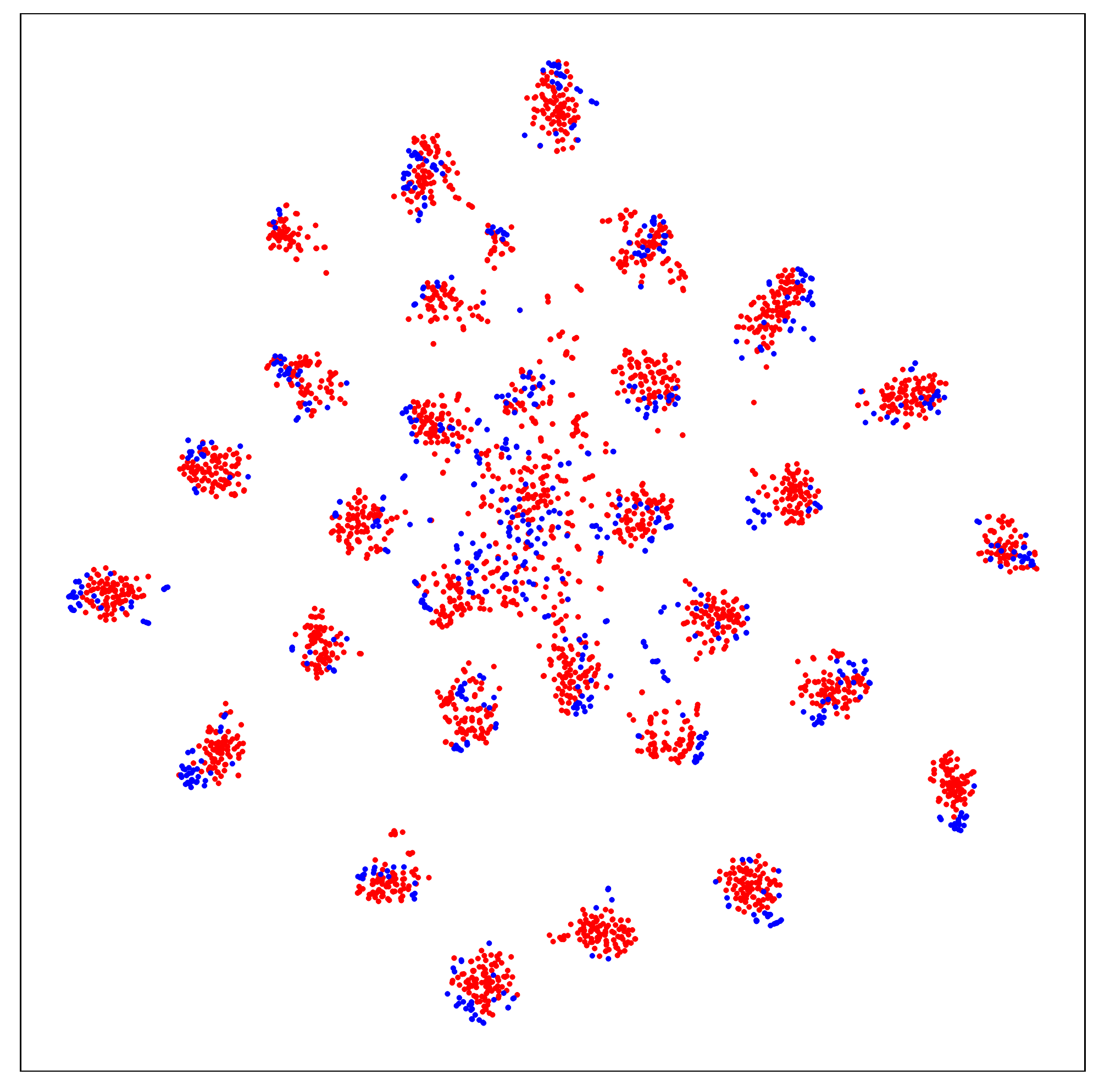}}
\\[1mm]
\subfloat[DALN]{\includegraphics[width=0.32\linewidth,height=0.08\textheight]{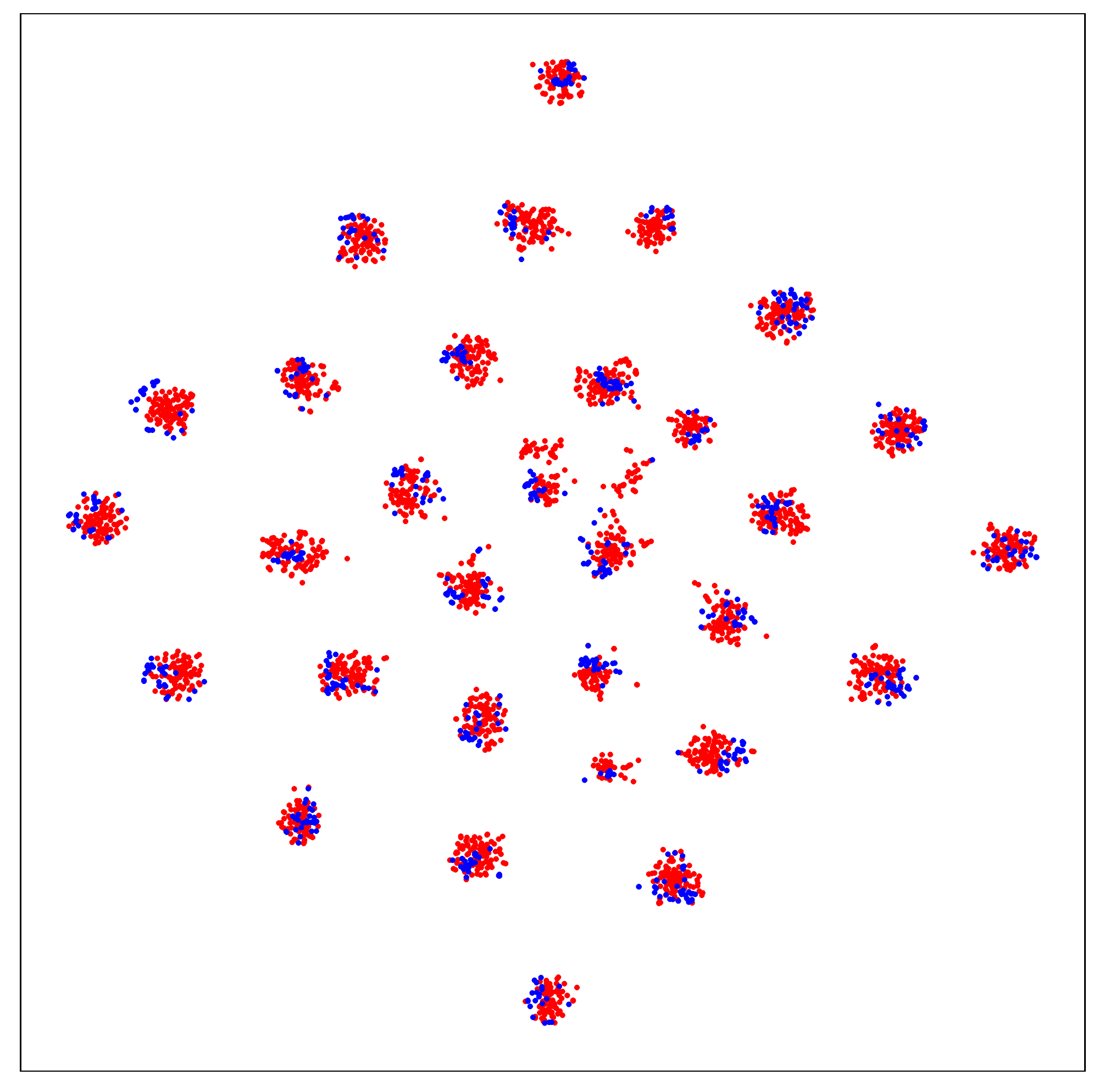}}
\subfloat[DANN+NWD]{\includegraphics[width=0.32\linewidth,height=0.08\textheight]{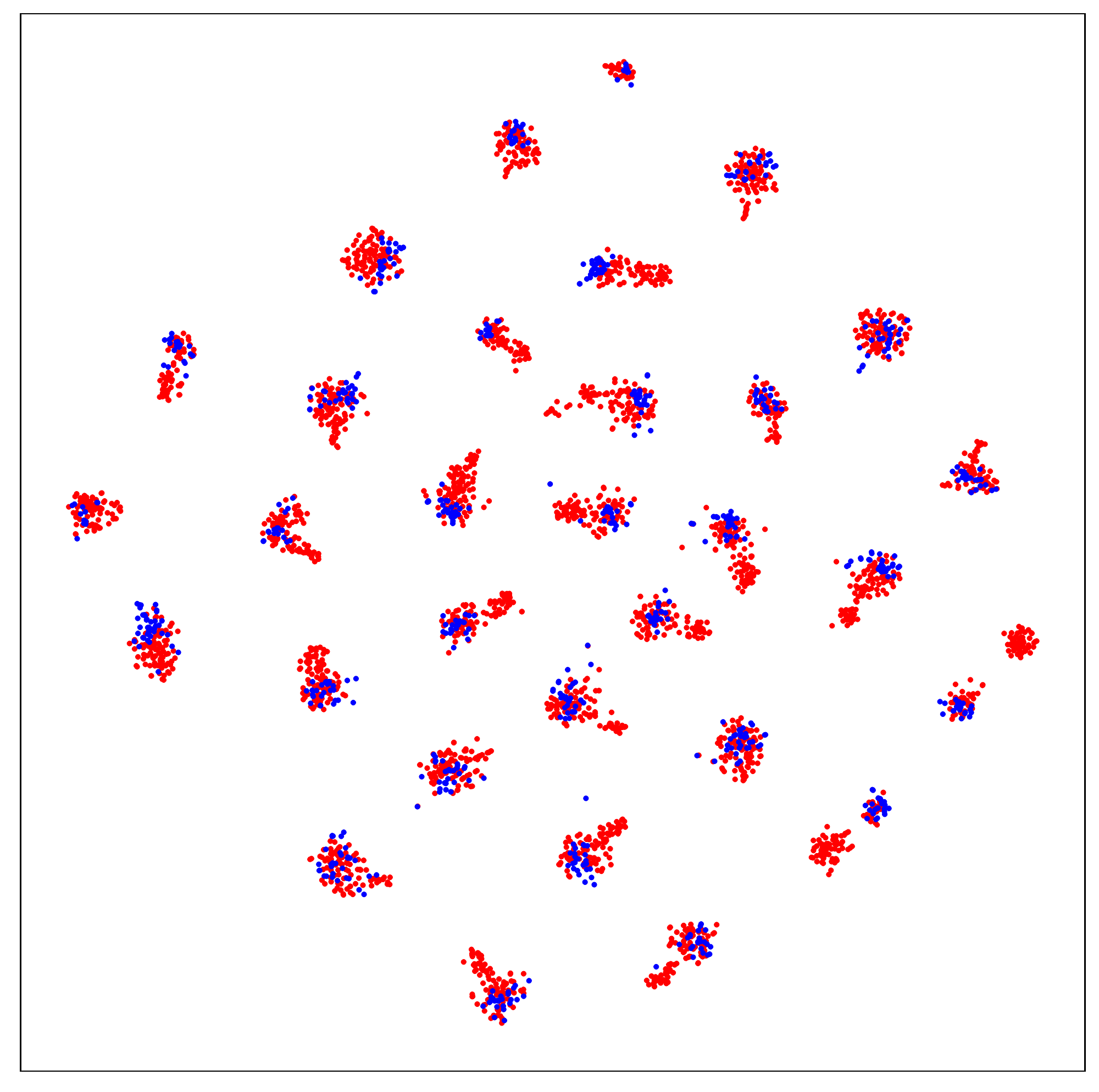}}
\subfloat[MDD+NWD]{\includegraphics[width=0.32\linewidth,height=0.08\textheight]{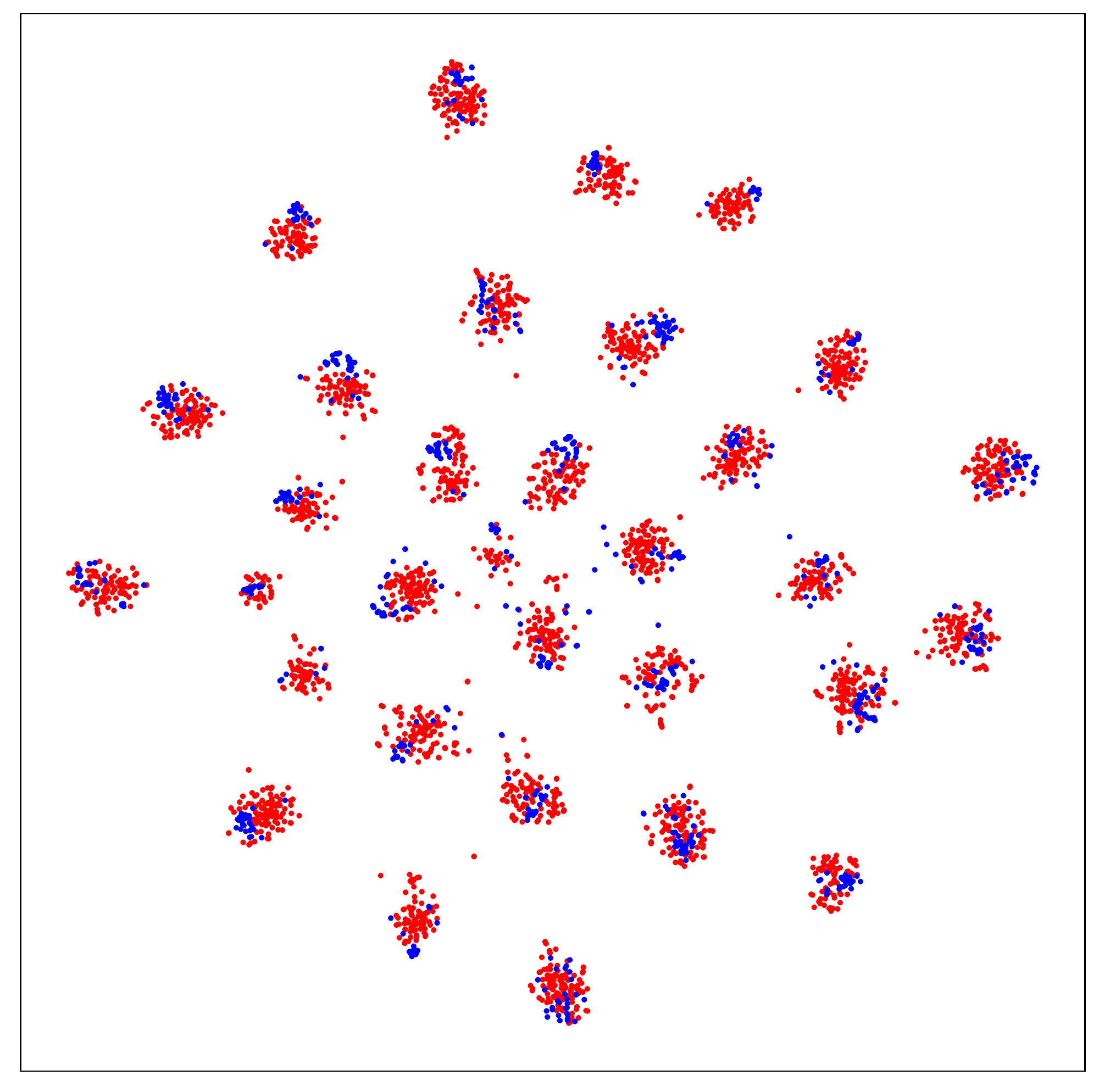}}
\caption{
t-SNE visualizations of feature distributions learned by different methods on task A→W of Office-31. Blue and red points represent source and target features, respectively. 
}
\vspace{-4mm}
\label{fig:tsne}
\end{figure}

\vspace{-2mm}
\noindent\textbf{t-SNE Visualization.} The feature representations of the ResNet-50, DANN, MDD, DALN, DANN+NWD, and MDD+NWD are visualized in \Fig{tsne} using t-SNE \cite{van2008visualizing}. Compared with the DANN and MDD, the proposed DALN not only confuses the feature representations, but also contributes to a more compact intra-class distribution and a more dispersed inter-class distribution, indicating that the features learned by the DALN are more discriminative. Combining the DANN and MDD with the proposed NWD, the intra-class features are pulled together while the inter-class features are pushed apart, demonstrating that the NWD can help them improve the discriminability.

\vspace{-2mm}


\section{Conclusions}
In this work, we present a simple yet effective adversarial paradigm, i.e., reusing the task-specific classifier as a discriminator. To achieve this paradigm, we designed a new discrepancy NWD that has definite guidance meaning and correspondingly built a discriminator-free adversarial UDA model, i.e., DALN, which learns transferable and discriminative representations while promising prediction determinacy and diversity. Moreover, we demonstrated that the proposed NWD can be used as a plug-and-play regularizer to the existing methods, which helps these methods achieve more competitive performance. Extensive experiments on a variety of datasets demonstrate the effectiveness and superiority of the proposed method.

\section{Acknowledgements}
This work was supported in part by the National Natural Science Foundation of China under Grant 61727809, in part by the Special Fund for Key Program of Science and Technology of Anhui Province under Grant 201903c08020002, and in part by the National Key Research and Development Program of China under Grant 2019YFC0117800.

{
    \clearpage
    \small
    \bibliographystyle{ieee_fullname}
    \bibliography{macros,main,supp}
    
}

\clearpage
\appendix
\setcounter{page}{1}

\twocolumn[
\centering
\Large
\textbf{Reusing the Task-specific Classifier as a Discriminator:\\
Discriminator-free Adversarial Domain Adaptation} \\
\vspace{0.5em}\textbf{Supplementary Material} \\
\vspace{1.0em}
] 
\appendix

This supplementary material provides more details that are not presented in the main paper due to space limitations. In the following sections, we first provide the proof of the implicitly constructed discriminator $D = {\left\| C \right\|_ * }$ satisfying the K-Lipschitz constraint. Then, we prove that the expected target risk can be bounded by the expected and empirical measures of the Nuclear-norm 1-Wasserstein discrepancy (NWD) on the source and target domains. Finally, more implementation details, experimental results, and insight analysis are presented, including the detailed comparisons on the VisDA-2017 dataset, the extra comparisons on the DomainNet \cite{peng2019moment} dataset, and analyses regarding toy experiments, Proxy $\cal A$-distance, self-correlation matrix, convergence, and trade-off parameters ($\lambda$ and $\gamma$).

\section{K-Lipschitz constraint}
To prove that the implicitly constructed discriminator $D = {\left\| C \right\|_ * }$ satisfies the K-Lipschitz constraint, where the classifier $C$ consists of a fully connected layer ${L_c}\left(  \cdot  \right)$ and a softmax function ${S_m}\left(  \cdot  \right)$, we first analyze ${L_c}\left(  \cdot  \right)$ and ${S_m}\left(  \cdot  \right)$, respectively.

\begin{definition}\label{def:1}
Given two metric spaces $\left( {M,{d_m}} \right)$ and $(N,{d_n})$, where $d_m$ denotes the metric on the compact set $M \subseteq {\mathbb R ^m}$ and $d_n$ is the metric on the compact set $N \subseteq {\mathbb R ^n}$, a function $h:M \to N$ is called K-Lipschitz continuous if there exists a real constant $K \ge 0$ (the minimum $K$ called Lipschitz constant) such that, for $\forall {m_1},{m_2} \in M$, the following holds
\begin{equation}\label{eq_supp:1}
    {\left\| h \right\|_L} = \mathop {\sup }\limits_{{m_1} \ne {m_2}} \frac{{{d_N}\left( {h\left( {{m_1}} \right),h\left( {{m_2}} \right)} \right)}}{{{d_M}\left( {{m_1},{m_2}} \right)}} \le K.
\end{equation}
\end{definition}

\begin{prop}\label{prop:1}
Given two metric spaces $\left( {F,\left|  \cdot  \right|} \right)$ and $\left( {O,\left|  \cdot  \right|} \right)$, where $F \subseteq{\mathbb R ^d}$ and $O \subseteq{\mathbb R ^k}$ denote the compact input feature set and output set, respectively, $\left|  \cdot  \right|$ denotes the Frobenius norm in $O$ or $F$. Then, for every input feature $f \in F$, the Lipschitz constant $K$ of the fully connected layer ${L_c}\left( f \right) = Wf + b$, where ${L_c}\left(  \cdot  \right):F \to O$ maps the feature $f \in F$ to the output $o \in O$, $W \in {\mathbb{R}^{k \times d}}$ denotes the weight matrix, and $b \in {\mathbb R ^k}$ denotes the bias vector, has a upper bound ${\left\| W \right\|_F}$.
\end{prop}

\begin{proof}
Given features ${f_1},{f_2} \in F$, if $f_1=f_2$, we have $\left| {h\left( {{f_1}} \right) - h\left( {{f_2}} \right)} \right| = K\left| {{f_1} - {f_2}} \right| = 0$, and if $f_1 \ne{f_2}$, we have
\begin{align}
    \left| {{L_c}\left( {{f_1}} \right) - {L_c}\left( {{f_2}} \right)} \right| &= \left| {\left( {W{f_1} + b} \right) - \left( {W{f_2} + b} \right)} \right| \notag\\
    &=\left| {W\left( {{f_1} - {f_2}} \right)} \right|.\notag
\end{align}
Meanwhile, the spectral norm of the matrix $W$ induced by $\left| f \right|$ is defined as
\begin{equation}
    {\left\| W \right\|_2} = \mathop {\max }\limits_{f \ne {\mathbf{0}}} \frac{{\left| {Wf} \right|}}{{\left| f \right|}} = {\sigma _{\max }}\notag,
\end{equation}
where ${\sigma _{\max }}$ is the maximum singular value obtained by singular value decomposition (SVD) on the matrix $W$. Therefore, according to \Def{1}, the Lipschitz constant $K$ is ${\left\| W \right\|_2}$. Additionally, the Frobenius norm of the matrix $W$ is defined as
\begin{equation}
    {\left\| W \right\|_F} = \sqrt {\sum\limits_{i = 1}^k {\sum\limits_{j = 1}^d {W_{i,j}^2} } }  = \sqrt {\sum\limits_{i = 1}^r {\sigma _i^2} }\notag,
\end{equation}
where $r = \min \left\{ {k,d} \right\}$, ${\sigma _i}$ denotes the i-th singular value.
Thus, for every ${f_1},{f_2} \in F$, we have
\begin{align}
    \left| {{L_c}\left( {{f_1}} \right) - {L_c}\left( {{f_2}} \right)} \right| &= \left| {W\left( {{f_1} - {f_2}} \right)} \right| \notag \\
    &\leqslant {\left\| W \right\|_2}\left| {{f_1} - {f_2}} \right| \notag \\
    &= K\left| {{f_1} - {f_2}} \right| \notag \\
    &\leqslant {\left\| W \right\|_F}\left| {{f_1} - {f_2}} \right|.\notag
\end{align}
\end{proof}

According to \Prop{1}, ${\left\| W \right\|_F}$ can be an upper bound of the Lipschitz constant $K$. As a widely used strategy, the weight decay (implemented with Frobenius norm regularization), which improves the generalization performance of a DNN model through minimizing an additional term ${\lambda _0}\left\| W \right\|_F^2$ (where $\lambda _0$ is a trade-off parameter), can simultaneously enforce the fully connected layer to satisfy the K-Lipschitz constraint.

\begin{definition}\label{def:2}
(Remark 4.6.10 in \cite{sohrab2003basic}) Given a function $h:M \to N$, where $M$ and $N$ denote the compact subsets of $\mathbb R^m$ and $\mathbb R^n$, $h$ will satisfy K-Lipschitz continuous if there exists a real constant $K \ge 0$ (the minimum $K$ called Lipschitz constant), such that, for any ${m_1},{m_2} \in M$, ${m_1}\ne{m_2}$, all the first partial derivatives are bounded by $K$.
\end{definition}

\begin{prop}
Given the compact output set $O \subseteq{\mathbb R^k}$ and prediction set $P \subseteq{\mathbb R^k}$, the softmax function ${S_m}\left(  \cdot  \right):O \to P$ mapping the output $o \in O$ to the prediction $p \in P$ satisfies the 1-Lipschitz constraint.
\end{prop}

\begin{proof}
Let $p = {S_m}\left( o \right)$, where $p_i$ is in the range from 0 to 1, for $\forall i \in 1 \ldots k$, then, we have
\begin{align}
    {p_i} = \frac{{{e^{{o_i}}}}}{{\sum\nolimits_{j = 1}^k {{e^{{o_j}}}} }} \quad &\forall i \in 1 \ldots k
    \notag \\
    \sum\limits_{i = 1}^k {{p_i}}  = 1 \quad &\forall i \in 1 \ldots k\notag.
\end{align}
For simplicity, we denote $\sum\nolimits_{j = 1}^k {{e^{{o_j}}}} $ as $\Sigma $. Then, the Jacobian matrix can be written as
\begin{equation}
    J = \left[ {\begin{array}{*{20}{c}}
  {\frac{{\partial {p_1}}}{{\partial {o_1}}}}& \cdots &{\frac{{\partial {p_1}}}{{\partial {o_k}}}} \\ 
   \vdots & \ddots & \vdots  \\ 
  {\frac{{\partial {p_k}}}{{\partial {o_1}}}}& \cdots &{\frac{{\partial {p_k}}}{{\partial {o_k}}}} 
\end{array}} \right].\notag
\end{equation}
According to the quotient rule, when $i=j$,  we have
\begin{align}
    \frac{{\partial {p_i}}}{{\partial {o_j}}} &= \frac{{{e^{{o_i}}}\Sigma  - {e^{{o_j}}}{e^{{o_i}}}}}{{{\Sigma ^2}}} \notag \\
    &= \frac{{{e^{{o_i}}}}}{\Sigma } \cdot \frac{{\Sigma  - {e^{{o_j}}}}}{\Sigma } \notag \\
    &= {p_i}\left( {1 - {p_j}} \right).\notag
\end{align}
When $i\ne{j}$, we have
\begin{align}
    \frac{{\partial {p_i}}}{{\partial {o_j}}} &= \frac{{0 - {e^{{o_j}}}{e^{{o_i}}}}}{{{\Sigma ^2}}} \notag \\
    &= - \frac{{{e^{{o_i}}}}}{\Sigma } \cdot \frac{{{e^{{o_j}}}}}{\Sigma } \notag \\
    &= - {p_i}{p_j}.\notag
\end{align}
Therefore, we have
\begin{equation}
    {J_{i,j}} = \left\{ {\begin{array}{*{20}{c}}
  {{p_i}\left( {1 - p{}_j} \right)}&{i = j} \\ 
  { - {p_i}{p_j}}&{i \ne j}.\notag
\end{array}} \right.
\end{equation}
Therefore, given $p_i$ in the range from 0 to 1, ${J_{i,j}}$ can be bounded by 1.
Thus, according to \Def{2}, softmax function satisfies the 1-Lipschitz constraint.
\end{proof}

\section{Generalization Bound}
In this section, we first provide the proof for \Lemma{1} and \Thm{1}. Then, based on \Thm{1}, we prove that the expected target risk can be bounded by the expected and empirical measures of NWD on the source and target domains.

\begin{lemma}\label{lemma:1}
(Lemma 1 of \cite{redko2017theoretical}; Lemma1 of \cite{shen2018wasserstein}) Let ${\nu_s},{\nu_t} \in \cal P \left( {\cal F} \right)$ denote the probability measures of the source and target domain features, $\rho \left( {{f^s},{f^t}} \right)$ be the cost of transporting a unit of material from location $f^s$ satisfying $f^s \sim {\nu_s}$ to location $f^t$ satisfying $f^t \sim {\nu_t}$, ${W_1}\left( {{\nu_s},{\nu_t}} \right)$ represent the NWD, and $K$ denote a Lipschitz constant. Given a family of classifiers $C \in {\cal H}_1$ and a ideal classifier ${C^ * } \in {{\cal H}_1}$ satisfying the K-Lipschitz constraint, where ${\cal H}_1$ is a subspace of $\cal H$, the following holds for every $C,{C^ * } \in {{\cal H}_1}$.
\begin{equation}\
    \left| {{\varepsilon _s}\left( {C,{C^ * }} \right) - {\varepsilon _t}\left( {C,{C^ * }} \right)} \right| \leqslant 2K{W_1}\left( {{\nu_s},{\nu_t}} \right).
\end{equation}
\end{lemma}

\begin{proof}
For every $C,{C^ * } \in {\mathcal{H}}_1$ satisfying the K-Lipschitz constraint, according to the triangle inequality, we have
\begin{align}
    {\left| {C\left( {{f^s}} \right) - {C^ * }\left( {{f^s}} \right)} \right|} \leqslant & \left| {C\left( {{f^s}} \right) - C\left( {{f^t}} \right)} \right| \notag \\
    &+ {\left| {C\left( {{f^t}} \right) - {C^ * }\left( {{f^s}} \right)} \right|} \notag \\
    \leqslant & \left| {C\left( {{f^s}} \right) - C\left( {{f^t}} \right)} \right| \notag \\
    &+ {\left| {C\left( {{f^t}} \right) - {C^ * }\left( {{f^t}} \right)} \right|} \notag \\
    &+ \left| {{C^ * }\left( {{f^s}} \right) - {C^ * }\left( {{f^t}} \right)} \right|. \notag
\end{align}
Therefore, for every ${f_1,f_2} \in \cal F$, the following holds,
\begin{align}
    \frac{{\left| {\left| {C\left( {{f^s}} \right) - {C^ * }\left( {{f^s}} \right)} \right| - \left| {C\left( {{f^t}} \right) - {C^ * }\left( {{f^t}} \right)} \right|} \right|}}{{\rho \left( {{f^s},{f^t}} \right)}} \leqslant 2K. \notag
\end{align}
For simplicity, we denote the discrepancy term $\left| {{\varepsilon _s}\left( {C,{C^ * }} \right) - {\varepsilon _t}\left( {C,{C^ * }} \right)} \right|$ as $dis$. Thus, given ${f_1,f_2} \in \cal F$, and $C'$ denotes the labeling function $C':\mathcal{F} \to \left[ {0,1} \right]$, for every $C,{C^ * } \in {{\cal H}_1}$, we have
\begin{flalign}
    \begin{split}
        \hspace{-0.5mm}
        dis&={\mathbb{E}}_{{\nu _t}}\left[ {\left| {C\left( {{f^t}} \right) - {C^ * }\left( {{f^t}} \right)} \right|} \right] - {\mathbb{E}_{{\nu _s}}}\left[ {\left| {C\left( {{f^s}} \right) - {C^ * }\left( {{f^s}} \right)} \right|} \right] \notag \\
        &\leqslant \mathop {\sup }\limits_{{{\left\| C' \right\|}_L} \leqslant K} {\mathbb{E}_{{\nu_s}}}\left[ {C'\left( {{f^s}} \right)} \right] - {\mathbb{E}_{{\nu_t}}}\left[ {C'\left( {{f^t}} \right)} \right] \notag \\
        &= 2K{W_1}\left( {{\nu_s},{\nu_t}} \right).
    \end{split}
\end{flalign}
\end{proof}

\begin{thm}\label{thm:1}
Based on \Lemma{1}, for every $C \in {\cal H}_1$, the following holds
\begin{equation}
    {\varepsilon _t}\left( C \right) \leqslant {\varepsilon _s}\left( C \right) + 2K{W_1}\left( {{\nu_s},{\nu_t}} \right) + {\eta ^ * },
\end{equation}
where ${\eta ^ * } = {\varepsilon _s}\left( {{C^ * }} \right) + {\varepsilon _t}\left( {{C^ * }} \right)$ is the risk of ideal joint hypothesis, which is a sufficiently small constant.
\end{thm}

\begin{proof}
Based on \Lemma{1}, we have
\begin{flalign}
    \begin{split}
        {\varepsilon _t}\left( C \right) &\leqslant {\varepsilon _t}\left( {{C^ * }} \right) + {\varepsilon _t}\left( {{C^ * },C} \right) \notag \\
        &={\varepsilon _t}\left( {{C^ * }} \right) + {\varepsilon _s}\left( {C,{C^ * }} \right) + {\varepsilon _t}\left( {{C^ * },C} \right) - {\varepsilon _s}\left( {C,{C^ * }} \right) \notag \\
        &={\varepsilon _t}\left( {{C^ * }} \right) + {\varepsilon _s}\left( {C,{C^ * }} \right) + {\varepsilon _t}\left( {C,{C^ * }} \right) - {\varepsilon _s}\left( {C,{C^ * }} \right) \notag \\
        &\leqslant {\varepsilon _t}\left( {{C^ * }} \right) + {\varepsilon _s}\left( {C,{C^ * }} \right) + 2K{W_1}\left( {{\nu_s},{\nu_t}} \right) \notag \\
        &\leqslant {\varepsilon _t}\left( {{C^ * }} \right) + {\varepsilon _s}\left( C \right) + {\varepsilon _s}\left( {{C^ * }} \right) + 2K{W_1}\left( {{\nu_s},{\nu_t}} \right) \notag \\
        &={\varepsilon _s}\left( C \right) + 2K{W_1}\left( {{\nu_s},{\nu_t}} \right) + {\eta ^ * }.
    \end{split}
\end{flalign}
\end{proof}

Therefore, the expected target risk can be bounded by the expected measures of the NWD on the source and target domain distributions. Furthermore, we show the convergence of the empirical measures to the expected measures of the NWD on the source and target domain samples.

\begin{definition}
Given $p \geqslant{1}$ and $\eta > 0$, a probability measure $\nu$ on $\cal F$ satisfies ${T_p}\left( \eta  \right)$ if the inequality
    \begin{equation}
        W\left( {\nu ',\nu} \right) \leqslant \sqrt {\frac{2}{\eta }H\left( {\nu'|\nu} \right)} ,
    \end{equation}
where 
\begin{equation}
    H\left( {\nu '|\nu } \right) = \int {\frac{{d\nu '}}{{d\nu }}} \log \frac{{d\nu '}}{{d\nu }}d\nu,
\end{equation}
holds for any probability measure $\nu'$.
\end{definition}

\begin{lemma} \label{lemma:2}
(Theorem 2.1 of \cite{bolley2007quantitative}; Theorem 1 of \cite{redko2017theoretical}; Theorem 2 of \cite{shen2018wasserstein}) Let $\nu \in \mathcal{P}\left( \mathcal{F} \right)$ be a probability measure in representation space $\cal F$, where $\cal F$ is a subspace of $\mathbb R^d$, satisfying ${T_1}\left( {{\eta ^ * }} \right)$ inequality. Let $\hat \nu  = \frac{1}{N}\sum\nolimits_{i = 1}^N {{\delta _{{f_i}}}}$ be its associated empirical measure defined on a sample set $\left\{ {{f_i}} \right\}_{i = 1}^N$ of size $N$ drawn i.i.d from $\nu$. Then for any $d' > d$ and $\eta ' < {\eta ^ * }$, there exists some constant $N_0$ depending on $d'$ and some square exponential moment of $\nu$ such that for any $\epsilon >0$ and $N \geqslant {N_0}\max \left( {{\epsilon ^{ - (d + 2)}},1} \right)$, the following holds
\begin{equation}
    \mathbb{P}\left[ {{W_1}\left( {\nu,\hat \nu} \right) > \epsilon } \right] \leqslant \exp \left( { - \frac{{\eta '}}{2}N{\epsilon ^2}} \right),
\end{equation}
where $d'$, $\eta '$ can be calculated explicitly.
\end{lemma}

\begin{thm}
(Theorem 3 of \cite{redko2017theoretical}; Theorem 3 of \cite{shen2018wasserstein}) Under the assumption of \Lemma{1} and \Lemma{2}, let two probability measures ${\nu _s, \nu _t} \in {\cal P \left( \cal F \right)}$ of the source and target domain features satisfy the ${T_1}\left( {{\eta ^ * }} \right)$ inequality, $F_s$ and $F_t$ be two sample sets of size $N_s$ and $N_t$ drawn i.i.d from $\nu _s$ and $\nu _t$, respectively. Let ${{\hat \nu }_s} = \frac{1}{{{N_s}}}\sum\nolimits_{i = 1}^{{N_s}} {{\delta _{f_i^s}}} $ and ${{\hat \nu }_t} = \frac{1}{{{N_t}}}\sum\nolimits_{i = 1}^{{N_t}} {{\delta _{f_i^t}}} $ be the associated empirical measures. Then for any $d' > d$ and $\eta ' < \eta ^*$, there exists some constant $N_0$ depending on $d'$ such that for any $\delta  > 0$ and $\min \left( {{N_s},{N_t}} \right) \geqslant {N_0}\max \left( {{\delta ^{ - (d' + 2)}},1} \right)$ with probability at least $1-\delta$ for all $C$, the following holds
\begin{align}
    {\varepsilon _t}\left( C \right) \leqslant& {\varepsilon _s}\left( C \right) + 2K{W_1}\left( {{{\hat \nu}_s},{{\hat \nu}_t}} \right) + {\eta ^ * } \notag \\
    &+ 2K\sqrt {{{2\log \left( {\frac{1}{\delta }} \right)} \mathord{\left/
 {\vphantom {{2\log \left( {\frac{1}{\delta }} \right)} {\eta '}}} \right.
 \kern-\nulldelimiterspace} {\eta '}}} \left( {\sqrt {\frac{1}{{{N_s}}}}  + \sqrt {\frac{1}{{{N_t}}}} } \right),
\end{align}
where ${\eta ^ * } = {\varepsilon _s}\left( {{C^ * }} \right) + {\varepsilon _t}\left( {{C^ * }} \right)$ is a sufficiently small constant representing the ideal combined risk.
\end{thm}

\begin{proof}
Based on \Lemma{1} and \Lemma{2}, we have
\begin{flalign}
    \begin{split}
        {\varepsilon _t}\left( C \right) \leqslant& {\varepsilon _s}\left( C \right) + 2K{W_1}\left( {{\nu_s},{\nu_t}} \right) + {\eta ^ * } \notag \\
        \leqslant& {\varepsilon _s}\left( C \right) + 2K{W_1}\left( {{\nu_s},{{\hat \nu}_s}} \right){\text{ + }}2K{W_1}\left( {{{\hat \nu}_s},{\nu_t}} \right){\text{ + }}{\eta ^ * } \notag \\
        \leqslant& {\varepsilon _s}\left( C \right) + 2K{W_1}\left( {{{\hat \nu}_s},{{\hat \nu}_t}} \right) + 2K{W_1}\left( {{{\hat \nu}_t},{\nu_t}} \right) + {\eta ^ * } \notag \\
        &+ 2K\sqrt {{{2\log \left( {\frac{1}{\delta }} \right)} \mathord{\left/
         {\vphantom {{2\log \left( {\frac{1}{\delta }} \right)} {{N_s}\eta '}}} \right.
         \kern-\nulldelimiterspace} {{N_s}\eta '}}} \notag \\
        \leqslant& {\varepsilon _s}\left( C \right){\text{ + }}2K{W_1}\left( {{{\hat \nu}_s},{{\hat \nu}_t}} \right) + {\eta ^ * }\notag \\
        &+ 2K\sqrt {{{2\log \left( {\frac{1}{\delta }} \right)} \mathord{\left/{\vphantom {{2\log \left( {\frac{1}{\delta }} \right)} {\eta '}}} \right.\kern-\nulldelimiterspace} {\eta '}}} \left( {\sqrt {\frac{1}{{{N_s}}}}  + \sqrt {\frac{1}{{{N_t}}}} } \right).
    \end{split}
\end{flalign}
\end{proof}

\section{Implementation details}
The proposed method is implemented based on the PyTorch framework running on a GPU (Tesla-V100 32 GB). Following the existing methods \cite{ganin2016domain,NEURIPS2018_ab88b157}, the ResNet50 or ResNet101 pretrained on the ImageNet is used as the feature extractor $G$, in which we use the bottleneck layer used in \cite{jin2020minimum} to replace the last fully connected layer. Classifier $C$ is a fully connected layer depending on the specific task . The setting of the gradient reverse layer follows that of \cite{ganin2016domain}. The SGD optimizer is used to train the model with a moment of 0.9, a weight decay of 1e-3, a batch size of 36, and a cropped image size of 224×224. The initial learning rate of classifier $C$ is set to 5e-3, which is 10 times larger than that of the feature extractor $G$. Additionally, to facilitate model training, we use the annealing strategy \cite{ganin2015unsupervised} for the decay of the learning rate. \textbf{One can refer to our provided code for more implementation details.}

\section{Detailed results on VisDA-2017}
The detailed results on VisDA-2017 are shown in \Table{vis}. The proposed DALN achieves an average accuracy of 80.6 \%, outperforming the existing SOTA methods. Combining the proposed NWD with other methods, the performances of these methods are substantially improved by 22.6\%, 7.5\%, 5.2\%, and 4.9\% for the DANN, CDAN, MDD, and MCC, respectively. In particular, the improvements are evidently exhibited on categories including bus, car, person, and truck. These results come from the proposed NWD helping these methods distinguish some confusing class pairs such as bus and car, and train and truck. 

\section{Extra experiments on DomainNet}
We further conduct an experiment on DomainNet \cite{peng2019moment} (containing 0.6 million images, 345 categories, and 6 sub-domains), which consists of 30 sub-experiments. And the batch size is 64 for DomainNet. As the results shown in \Table{domainnet}, DALN outperforms the previous SOTA methods impressively in terms of the average accuracy. Such encouraging results demonstrate the superiority of DALN for processing complex datasets.

\begin{table*}
\centering
\caption{Classification accuracy (\%) on VisDA-2017  for unsupervised domain adaptation (using ResNet-101 as the backbone). $^\dag$ denotes that the results are reproduced using the publicly released  code. The best accuracy is indicated in \textbf{\textcolor{red}{bold red}} and the second best accuracy is indicated in \textcolor{blue}{{\ul undelined blue}}.}
\resizebox{0.99\textwidth}{!}{
\begin{tabular}{lccccccccccccc}
\hline
Method&plane&bcycl&bus&car&horse&knife&mcycl&person&plan&sktbrd&train&truck&Avg\\\hline
ResNet-101\cite{he2016deep}&55.1&53.3&61.9&59.1&80.6&17.9&79.7&31.2&81.0&26.5&73.5&8.5&52.4\\
WDGRL$^\dag$\cite{shen2018wasserstein}&85.4&54.2&76.2&41.4&68.9&56.8&86.9&48.2&57.2&51.9&81.8&27.2&61.3\\
MCD\cite{saito2018maximum}&87.0&60.9&83.7&64.0&88.9&79.6&84.7&76.9&88.6&40.3&83.0&25.8&71.9\\
BSP\cite{chen2019transferability}&92.4&61.0&81.0&57.5&89.0&80.6&{\color[HTML]{0000FF}{\ul90.1}}&77.0&84.2&77.9&82.1&38.4&75.9\\
SWD\cite{lee2019sliced}&90.8&82.5&81.7&70.5&91.7&69.5&86.3&77.5&87.4&63.6&85.6&29.2&76.4\\
BNM\cite{cui2020towards}&89.6&61.5&76.9&55.0&89.3&69.1&81.3&65.5&90.0&47.3&89.1&30.1&70.4\\
GVB-GD$^\dag$\cite{cui2020gradually}&90.1&68.7&81.9&61.7&91.2&67.3&{\color[HTML]{FE0000}\textbf{90.2}}&76.5&90.2&77.8&{\color[HTML]{FE0000}\textbf{90.3}}&41.0&77.2\\
DADA\cite{tang2020discriminative}&92.9&74.2&82.5&65.0&90.9&93.8&87.2&74.2&89.9&71.5&86.5&{\color[HTML]{0000FF}{\ul48.7}}&79.8\\
TSA\cite{li2021transferable}&-&-&-&-&-&-&-&-&-&-&-&-&78.6\\
SCDA$^\dag$\cite{li2021semantic}&93.1&{\color[HTML]{0000FF}{\ul84.6}}&78.2&52.2&90.8&95.2&81.0&77.2&{\color[HTML]{0000FF}{\ul91.1}}&80.5&{\color[HTML]{0000FF}{\ul89.1}}&43.5&79.7\\\hline
\textbf{DALN(Ours)}&{\color[HTML]{0000FF}{\ul96.0}}&{\color[HTML]{FE0000}\textbf{86.3}}&74.3&50.0&{\color[HTML]{0000FF}{\ul92.4}}&94.7&83.5&76.4&91.0&87.2&88.4&47.4&80.6\\\hline
DANN\cite{ganin2016domain}&81.9&77.7&82.8&44.3&81.2&29.5&65.1&28.6&51.9&54.6&82.8&7.8&57.4\\
DANN+\textbf{NWD}&{\color[HTML]{0000FF}{\ul96.0}}&73.6&{\color[HTML]{0000FF}{\ul84.3}}&48.3&88.0&92.8&89.4&{\color[HTML]{333333}78.2}&89.1&{\color[HTML]{0000FF}{\ul90.9}}&88.7&40.3&80.0\\\hline
CDAN\cite{NEURIPS2018_ab88b157}&85.2&66.9&83.0&50.8&84.2&74.9&88.1&74.5&83.4&76.0&81.9&38.0&73.9\\
CDAN+\textbf{NWD}&94.8&80.0&84.2&56.0&92.3&91.5&{\color[HTML]{0000FF}{\ul90.1}}&{\color[HTML]{0000FF}{\ul78.7}}&88.0&{\color[HTML]{FE0000}\textbf{91.1}}&88.9&41.4&81.4\\\hline
MDD$^\dag$\cite{zhang2019bridging}&80.1&61.3&83.7&51.8&90.7&83.8&89.7&77.3&90.2&86.6&82.2&44.5&76.8\\
MDD+\textbf{NWD}&94.0&81.0&{\color[HTML]{FE0000}\textbf{86.2}}&63.5&90.5&{\color[HTML]{FE0000}\textbf{97.0}}&87.5&76.3&88.6&86.5&85.2&48.2&{\color[HTML]{0000FF}{\ul82.0}}\\\hline
MCC\cite{jin2020minimum}&88.1&80.3&80.5&{\color[HTML]{FE0000}\textbf{71.5}}&90.1&93.2&85.0&71.6&89.4&73.8&85.0&36.9&78.8\\
MCC+\textbf{NWD}&{\color[HTML]{FE0000}\textbf{96.1}}&82.7&76.8&{\color[HTML]{0000FF}{\ul71.4}}&{\color[HTML]{FE0000}\textbf{92.5}}&{\color[HTML]{0000FF}{\ul96.8}}&88.2&{\color[HTML]{FE0000}\textbf{81.3}}&{\color[HTML]{FE0000}\textbf{92.2}}&{\color[HTML]{333333}88.7}&84.1&{\color[HTML]{FE0000}\textbf{53.7}}&{\color[HTML]{FE0000}\textbf{83.7}}\\\hline
\end{tabular}\label{tab:vis}
}
\end{table*}
\begin{figure*}
\centering
\subfloat[Normal: Source only]{\includegraphics[width=0.24\textwidth]{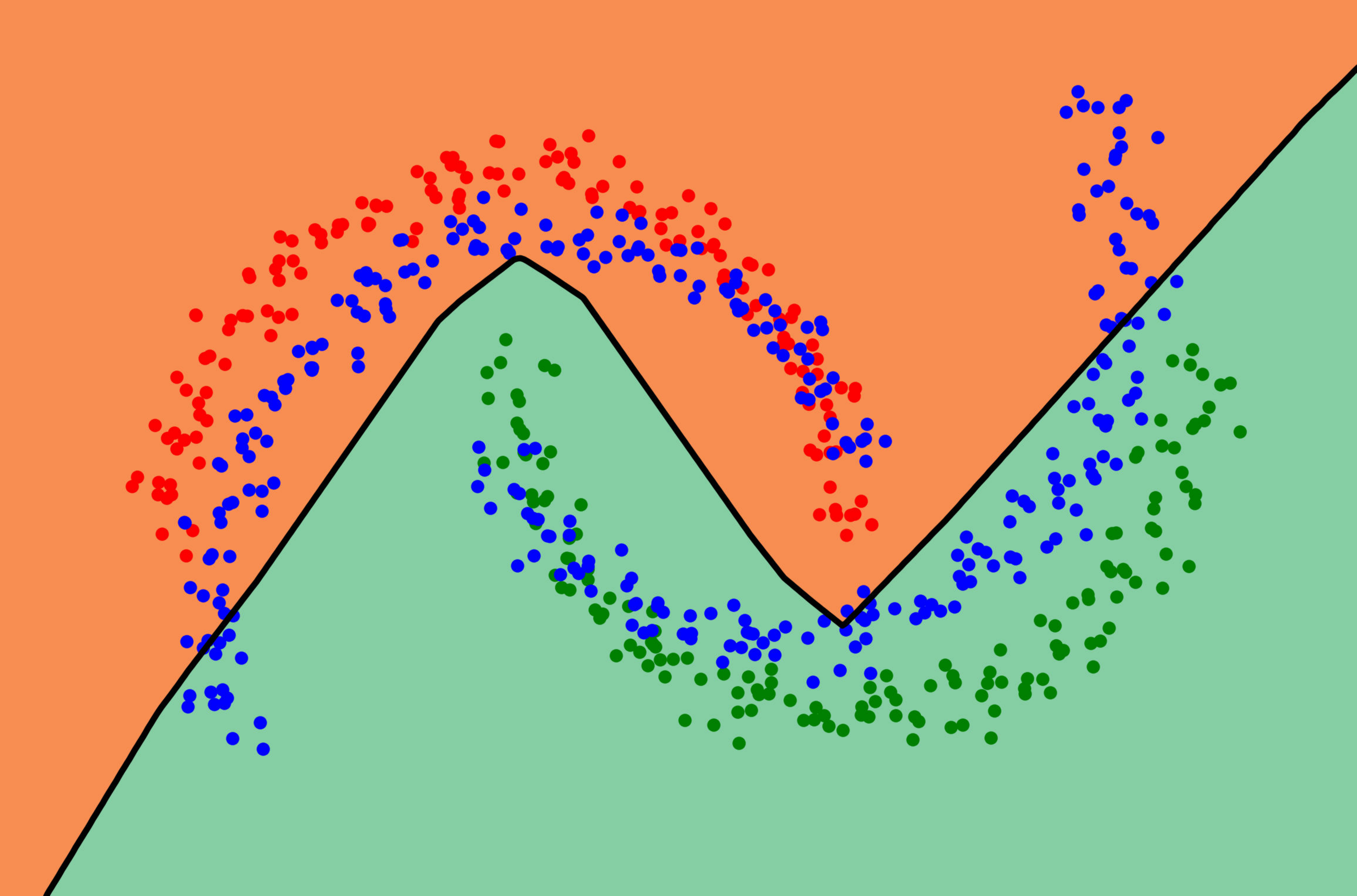}}\hspace{1mm}
\subfloat[Normal: DANN]{\includegraphics[width=0.24\textwidth]{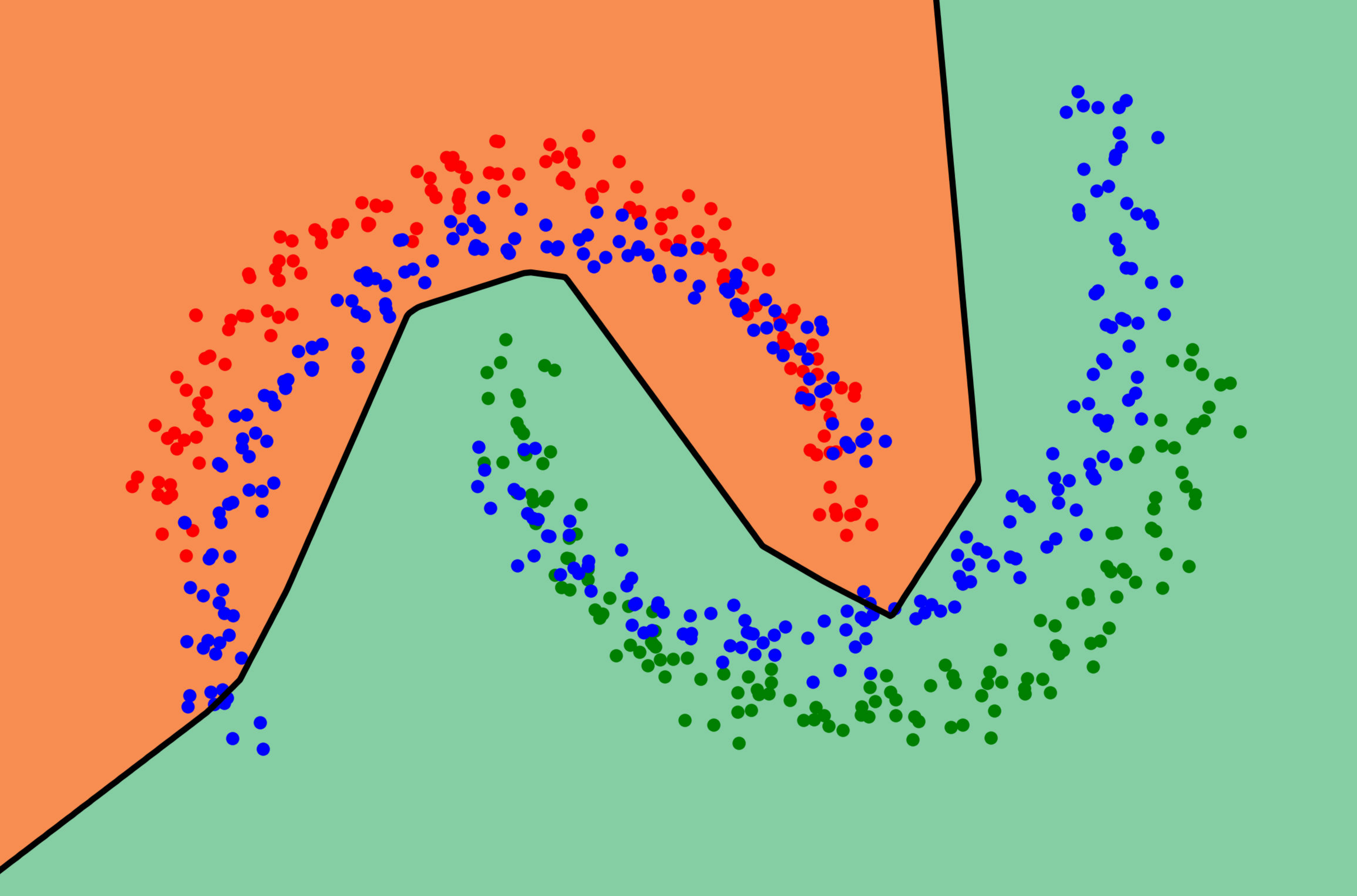}}\hspace{1mm}
\subfloat[Normal: MDD]{\includegraphics[width=0.24\textwidth]{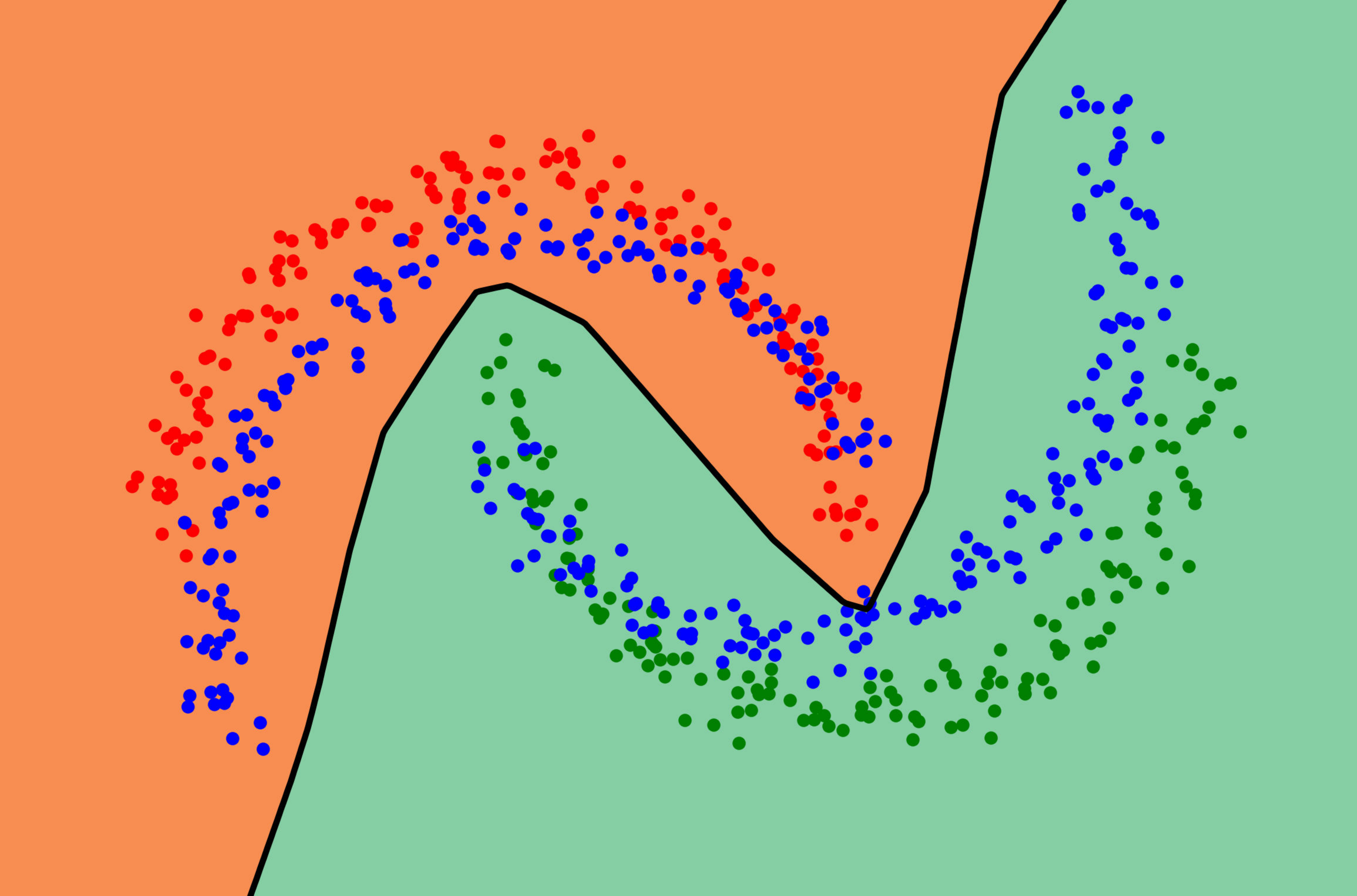}}\hspace{1mm}
\subfloat[Normal: DALN]{\includegraphics[width=0.24\textwidth]{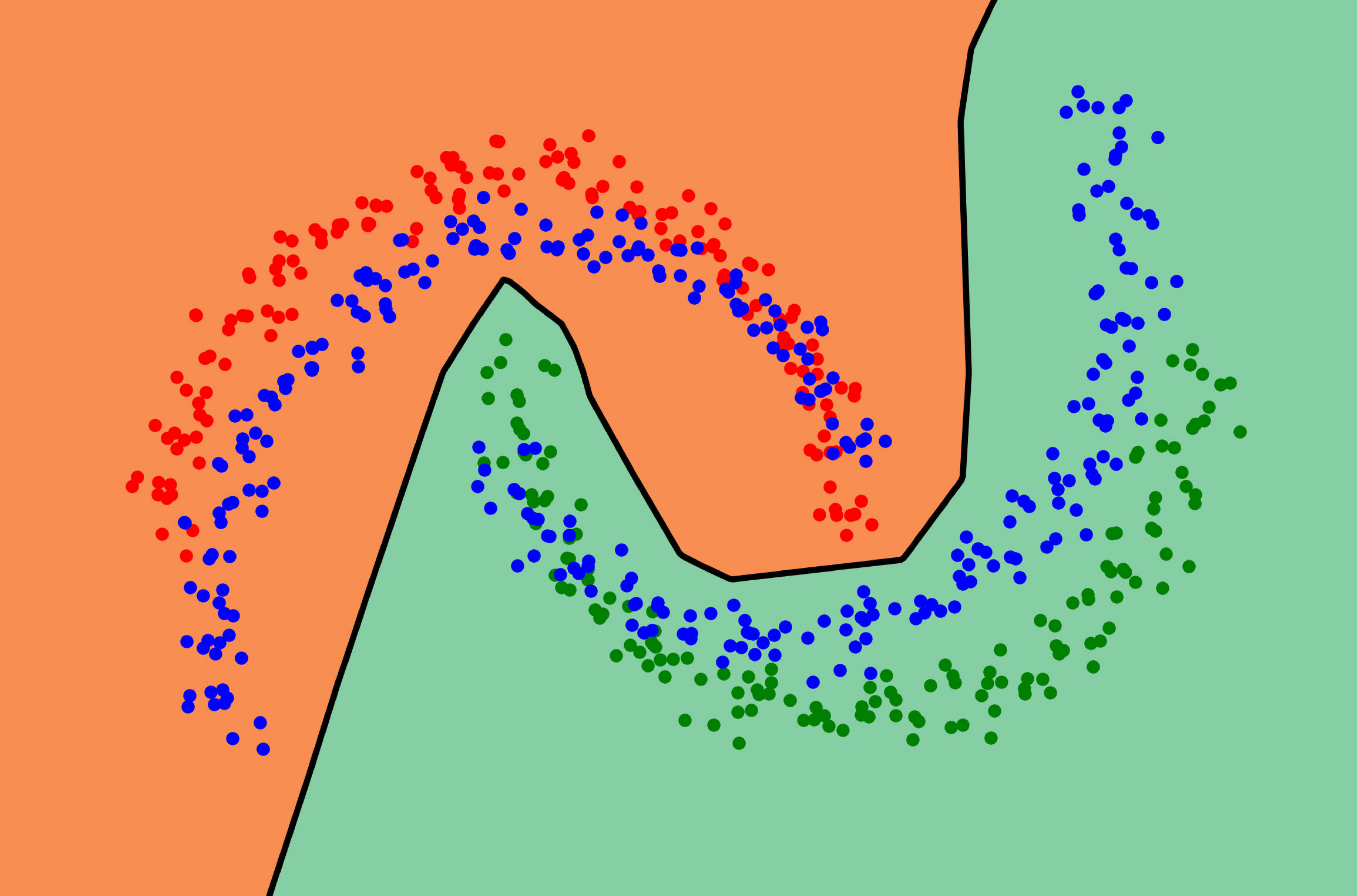}}\\
\subfloat[Sparse: Source only]{\includegraphics[width=0.24\textwidth]{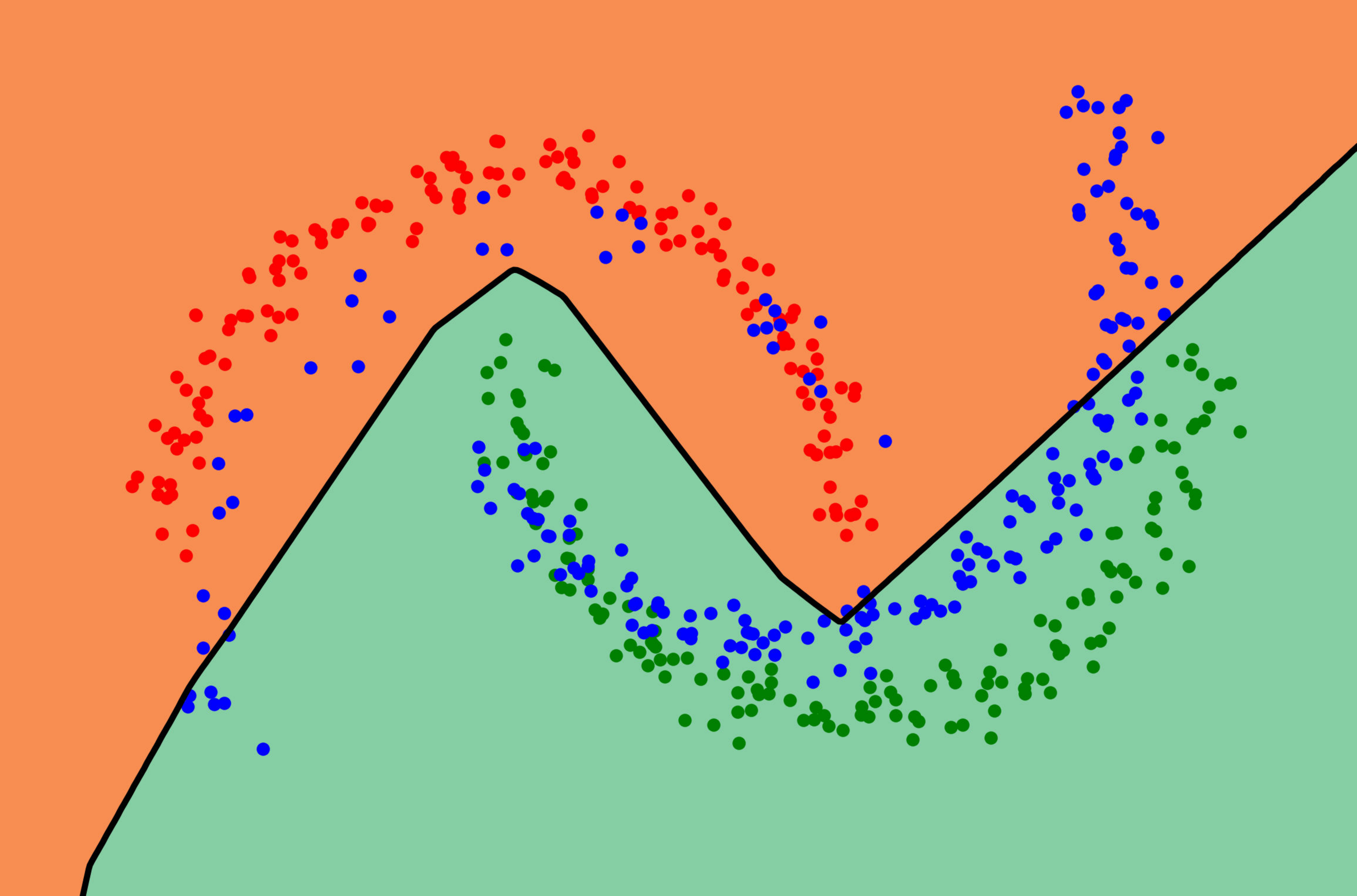}}\hspace{1mm}
\subfloat[Sparse: DANN]{\includegraphics[width=0.24\textwidth]{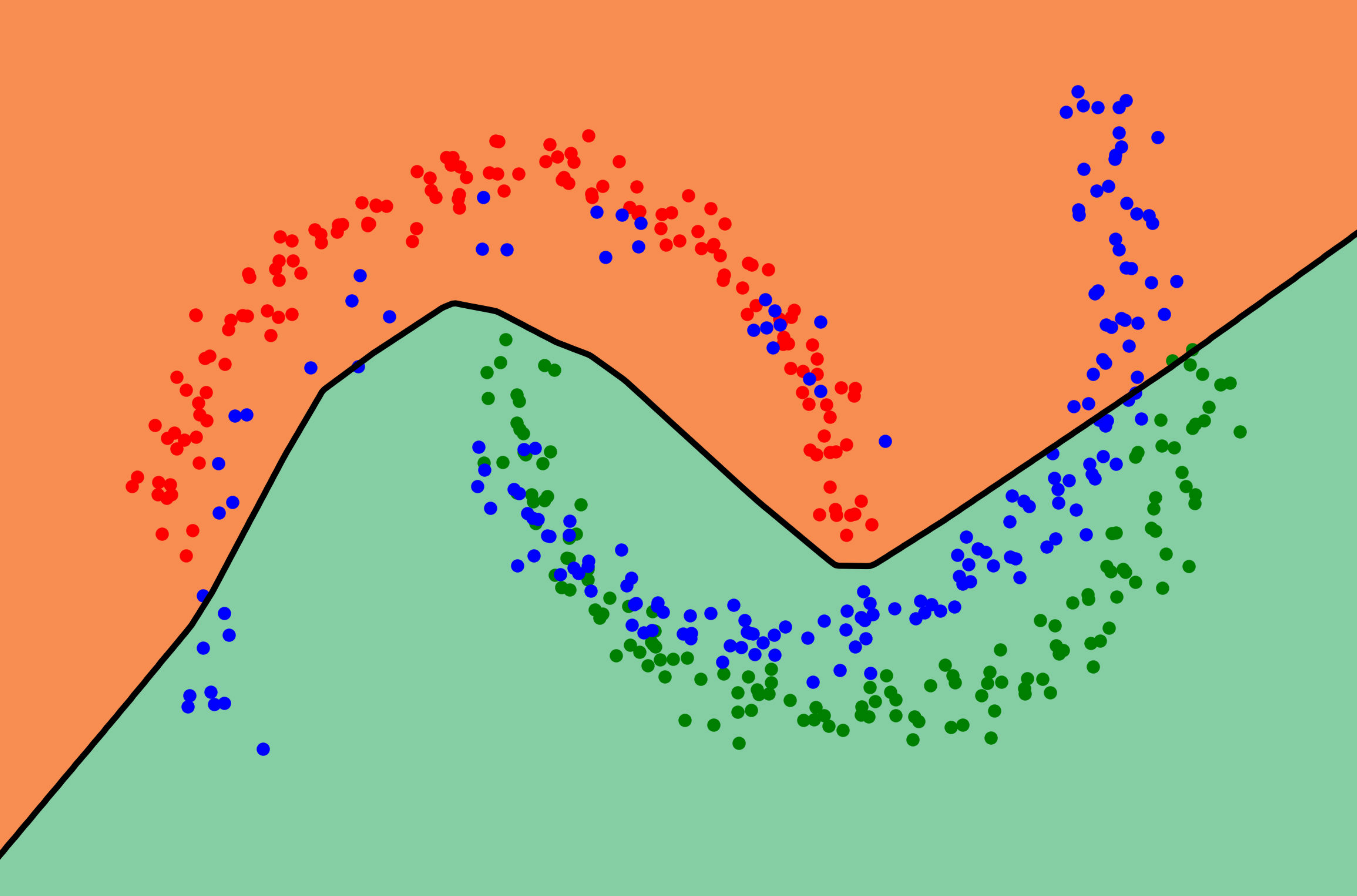}}\hspace{1mm}
\subfloat[Sparse: MDD]{\includegraphics[width=0.24\textwidth]{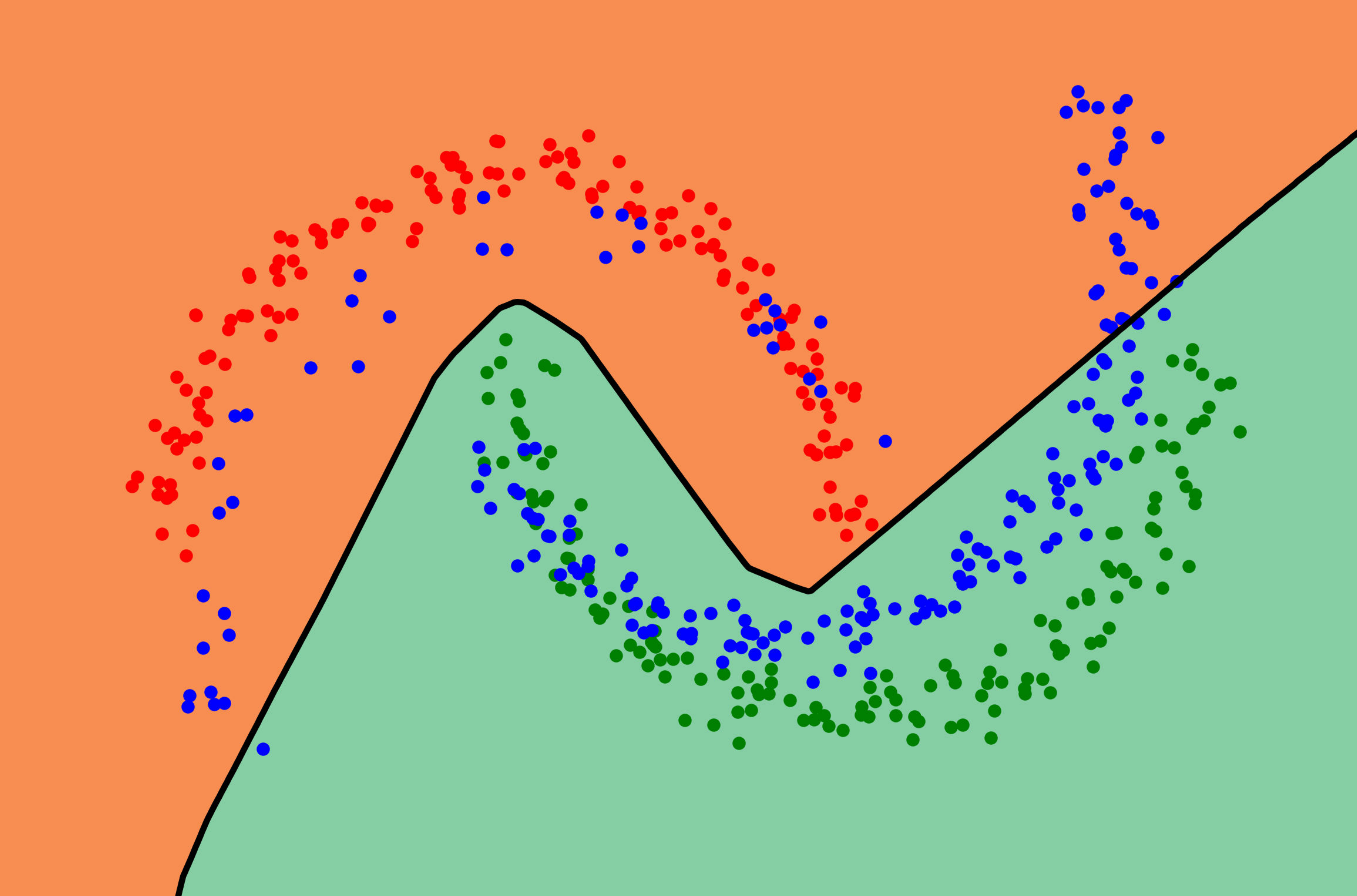}}\hspace{1mm}
\subfloat[Sparse: DALN]{\includegraphics[width=0.24\textwidth]{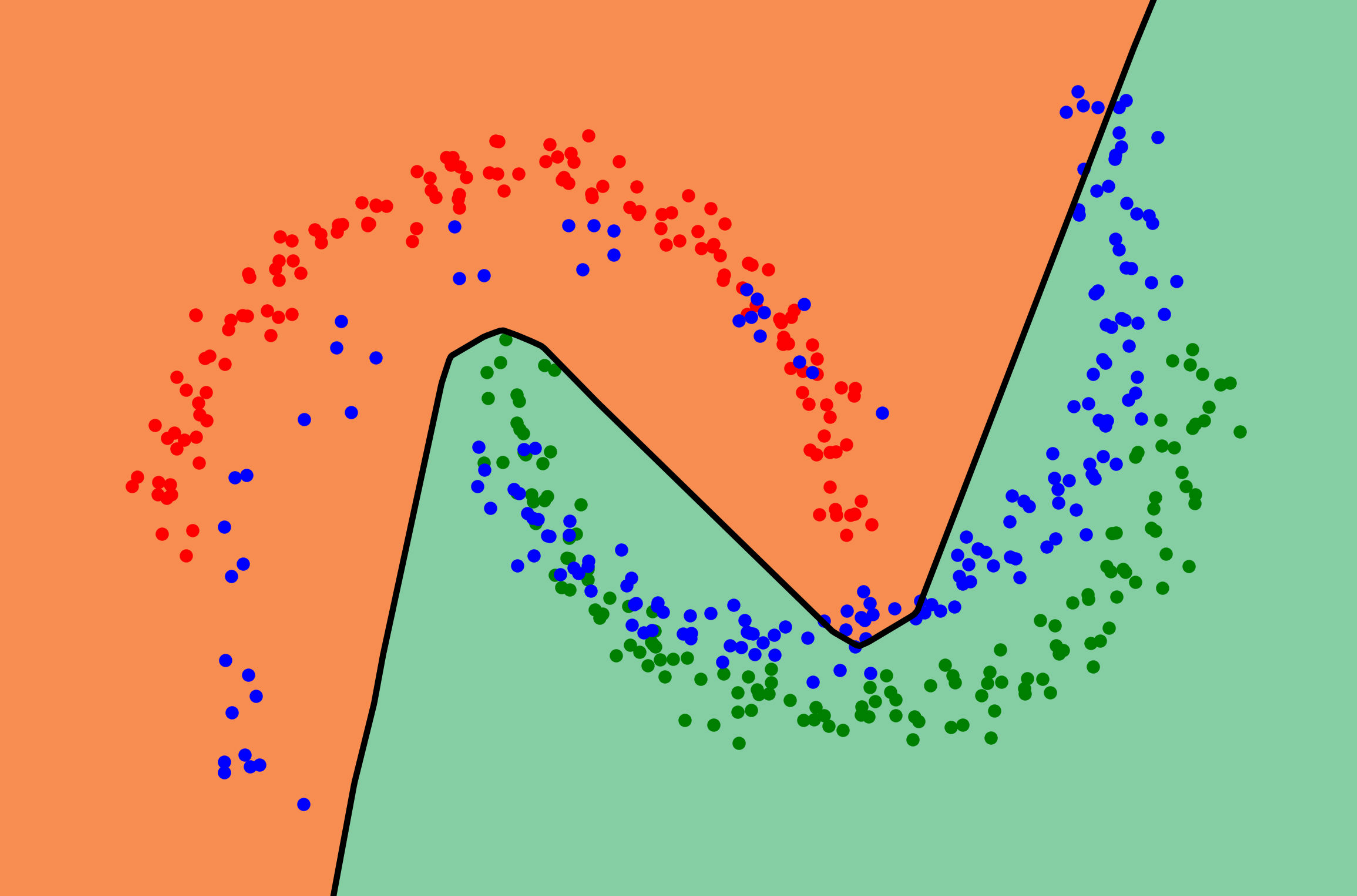}}
\vspace{-1mm}
\caption{
Comparisons of decision boundaries on a toy example dataset. Red points and green points denote classes 0 and 1 of source data, respectively. Blue points are target data generated  by rotating the source data distribution by 30 degrees. In the second row, we reduce the number of samples to 1/4 for the upper moon of the target domain via a sampling strategy, which thus generates imbalanced class samples. The orange and green regions are classified as classes 0 and 1 by the final decision boundary, respectively.
}
\label{fig:moon}
\end{figure*}

\section{Insight Analysis}
\noindent\textbf{Toy examples.} We perform toy experiments to discuss the inter twinning moons 2D problem \cite{pedregosa2011scikit}, helping analyze the learned decision boundary. The presented examples consider two cases. One case is generating balanced samples for both the source and target domains, and the other studies the class imbalance problem by reducing the samples of one class in the target domain. Specifically, for the first case, we generate 300 samples for two classes of the source samples labeled 0 and 1, and each class has 150 samples. As shown in \Fig{moon}, samples corresponding to 0 are denoted by an upper moon, while samples corresponding to 1 are denoted by a lower moon. Then, the target samples are generated by rotating the data distribution of the source samples by 30 degrees, resulting in a domain shift for the target domain. In this case, the number of samples of each class in the two domains are equal. In contrast, for the second case, we reduce the number of samples to 38 for the upper moon of the target domain via a sampling strategy, which thus generates imbalanced class samples. As shown in \Fig{moon}, for the first case, the model trained on the source-only data can correctly classify the source samples, but cannot perform properly in the overall target samples. DANN improves the decision boundary, but some samples in the upper moon are misclassified. MDD and DALN successfully classify both the source and target samples, but the proposed DALN achieves better classification performances compared with MDD. For the second case, except our DALN, both the DANN and MDD cannot learn a favorable decision boundary for the target samples. Some samples in the upper moon and lower moon are misclassified.
\begin{table}[htbp]
  \setlength{\abovecaptionskip}{0.cm}
  \setlength{\belowcaptionskip}{0.cm}
  \caption{Accuracy(\%) on DomainNet for UDA. In each sub-table, the column-wise domains are selected as the source domain and the row-wise domains are selected as the target domain.}
 \label{tab:domainnet}
 \centering
 \resizebox{\linewidth}{!}{
 \setlength{\tabcolsep}{0.5mm}{
   \begin{tabular}{|c|ccccccc||c|ccccccc|}
   \hline
   \textbf{ResNet-101} \cite{he2016deep}  & clp   & inf   & pnt   & qdr   & rel   & skt   & Avg.  & 
   \textbf{SCDA(21)} \cite{li2021semantic}  & clp   & inf   & pnt   & qdr   & rel   & skt   & Avg. \\
   \hline
   \hline
   clp & - & 19.3 & 37.5 & 11.1 & 52.2 & 41.0 & 32.2 & 
   clp & - & 18.6 & 39.3 & 5.1  & 55.0 & 44.1 & 32.4 \\
   
   inf & 30.2 & - & 31.2 & 3.6 & 44.0 & 27.9 & 27.4  & 
   inf & 29.6 & - & 34.0 & 1.4 & 46.3 & 25.4 & 27.3   \\
   
   pnt & 39.6 & 18.7 & - & 4.9 & 54.5 & 36.3 & 30.8  & 
   pnt & 44.1 & 19.0 & - & 2.6 & 56.2 & 42.0 & 32.8  \\
   
   qdr  & 7.0  & 0.9  & 1.4  & - & 4.1  & 8.3  & 4.3 & 
   qdr  & 30.0 & 4.9  & 15.0 & - & 25.4 & 19.8 & 19.0  \\
   
   rel  & 48.4 & 22.2 & 49.4 & 6.4 & - & 38.8 & 33.0 & 
   rel  & 54.0 & 22.5 & 51.9 & 2.3 & - & 42.5 & 34.6\\
   
   skt  & 46.9 & 15.4 & 37.0 & 10.9 & 47.0 & - & 31.4 &
   skt  & 55.6 & 18.5 & 44.7 & 6.4  & 53.2 & - & 35.7 \\
   
   Avg. & 34.4 & 15.3 & 31.3 & 7.4  & 40.4 & 30.5 & 26.6 & 
   Avg. & 42.6 & 16.7 & 37.0 & 3.6  & 47.2 & 34.8 & 30.3\\
   \hline
   
   \hline
   \textbf{BCDM(21)} \cite{Li21BCDM} & clp   & inf   & pnt   & qdr   & rel   & skt   & Avg.  & 
   \textbf{DALN(Ours)}  & clp   & inf   & pnt   & qdr   & rel   & skt   & Avg. \\
   \hline
   \hline
   clp & - & 19.9 & 38.5 & 15.1 & 53.2 & 43.9 & 34.1 & 
   clp & - & 20.0 & 40.2 & 11.4 & 57.5 & 45.4 & 34.9 \\
   
   inf & 31.9 & - & 32.7 & 6.9 & 44.7 & 28.5 & 28.9  & 
   inf & 35.2 & - & 34.7 & 4.7 & 47.9 & 29.0 & 30.3   \\
   
   pnt & 42.5 & 19.8 & - & 7.9 & 54.5 & 38.5 & 32.6  & 
   pnt & 45.3 & 19.2 & - & 3.2 & 57.4 & 40.0 & 33.0  \\
   
   qdr & 23.0 & 4.0 & 9.5 & - & 16.9 & 16.2 & 13.9   & 
   qdr & 27.5 & 4.2 & 13.2 & - & 21.8 & 16.6 & 16.7  \\
   
   rel & 51.9 & 24.9 & 51.2 & 8.7 & - & 40.6 & 35.5  & 
   rel & 55.6 & 22.8 & 54.0 & 5.1 & - & 40.4 & 35.6\\
   
   skt & 53.7 & 20.5 & 46.0 & 13.1 & 53.4 & - & 37.1 &
   skt & 59.0 & 19.9 & 46.0 & 8.3  & 56.3 & - & 37.9 \\
   
   Avg. & 40.6 & 17.8 & 35.6 & 10.3 & 44.3 & 33.5 & 30.4 & 
   Avg. & 44.5 & 17.2 & 37.6 & 6.5  & 48.2 & 34.3 & \textbf{31.4}\\
   \hline
   \end{tabular}
   }}
 \vspace{-1mm}
\end{table}

\begin{figure}[h]
\centering
\includegraphics[width=0.7\linewidth]{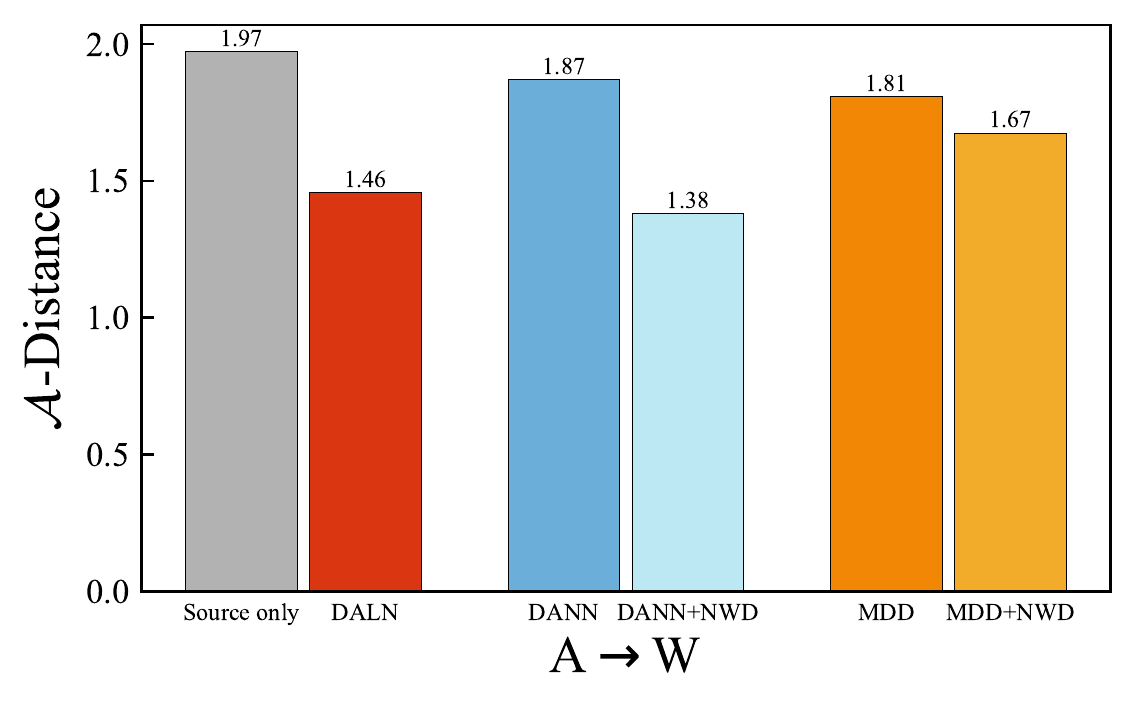}
\caption{
Visualization of the proxy $\cal A$-distance on task A→W of Office-31. Note that a smaller proxy $\cal A$-distance denotes better transferability.
}
\label{fig:adistance}
\end{figure}

\noindent\textbf{Proxy $\cal A$-distance.} As shown in \Fig{adistance}, we calculate the proxy $\cal A$-distance of the feature representations achieved by different methods based on task A→W of Office-31. Note that a smaller proxy $\cal A$-distance denotes better transferability. The proposed DALN achieves the lowest proxy $\cal A$-distance, demonstrating its superiority in learning transferable features. Moreover, by taking the NWD as a regularizer for DANN and MDD, their proxy $\cal A$-distances are considerably decreased, demonstrating the effectiveness of the proposed NWD in improving the transferability of the features.

\begin{figure}[htbp]
\centering
\subfloat[Source only]{\includegraphics[width=0.49\linewidth]{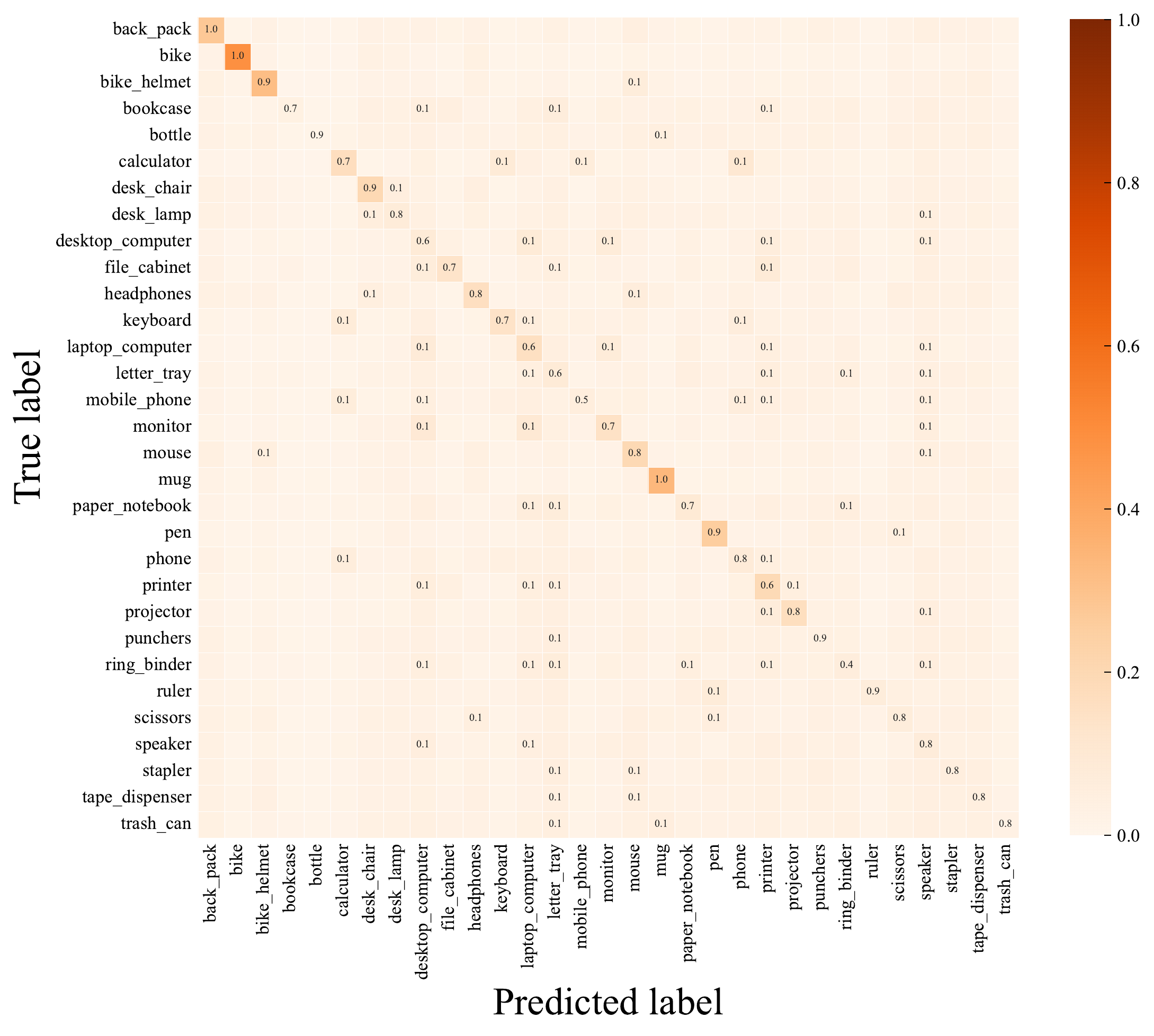}}
\subfloat[DALN]{\includegraphics[width=0.49\linewidth]{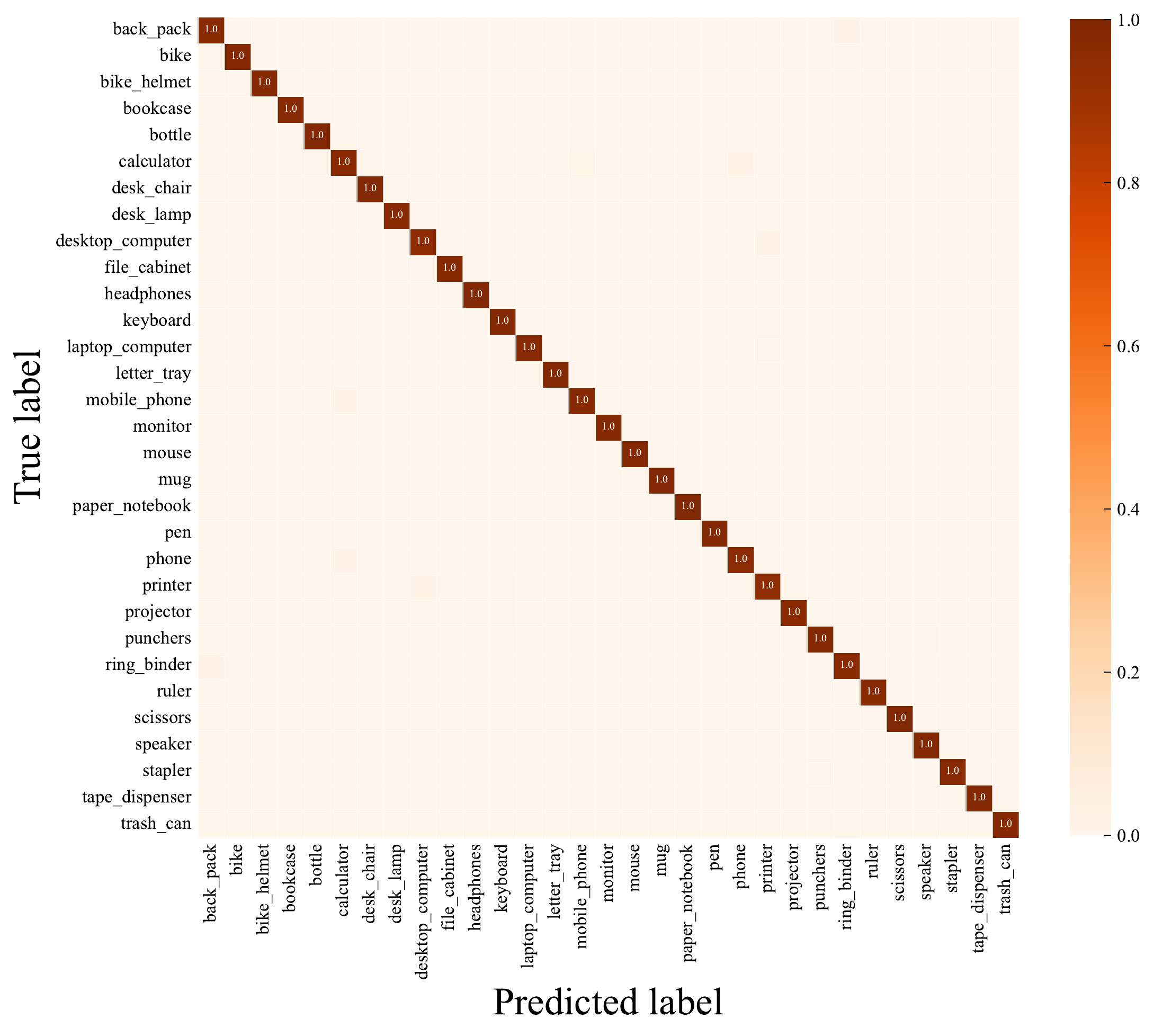}}
\caption{
The self-correlation matrices of the predictions on the target domain on task A→W of Office-31. (Zoom in for a clear visualization.)
}
\label{fig:self_correlation}
\end{figure}

\noindent \textbf{Self-correlation matrix.} As shown in \Fig{self_correlation}, the model trained on the source-only data generates large values on the off-diagonal elements for the target domain samples. In contrast, with the adaptation of the proposed paradigm, the values of the self-correlation matrix generated from the target samples are highly concentrated on the main diagonal as shown in \Fig{self_correlation}(b). Thus, the intra-class correlation $I_a$ is increased and the inter-class correlation $I_e$ is decreased, which demonstrates the effectiveness of the proposed method.

\begin{figure}[htbp]
\centering
\vspace{-4mm}
\subfloat[A→W]{
    \includegraphics[width=0.48\linewidth]{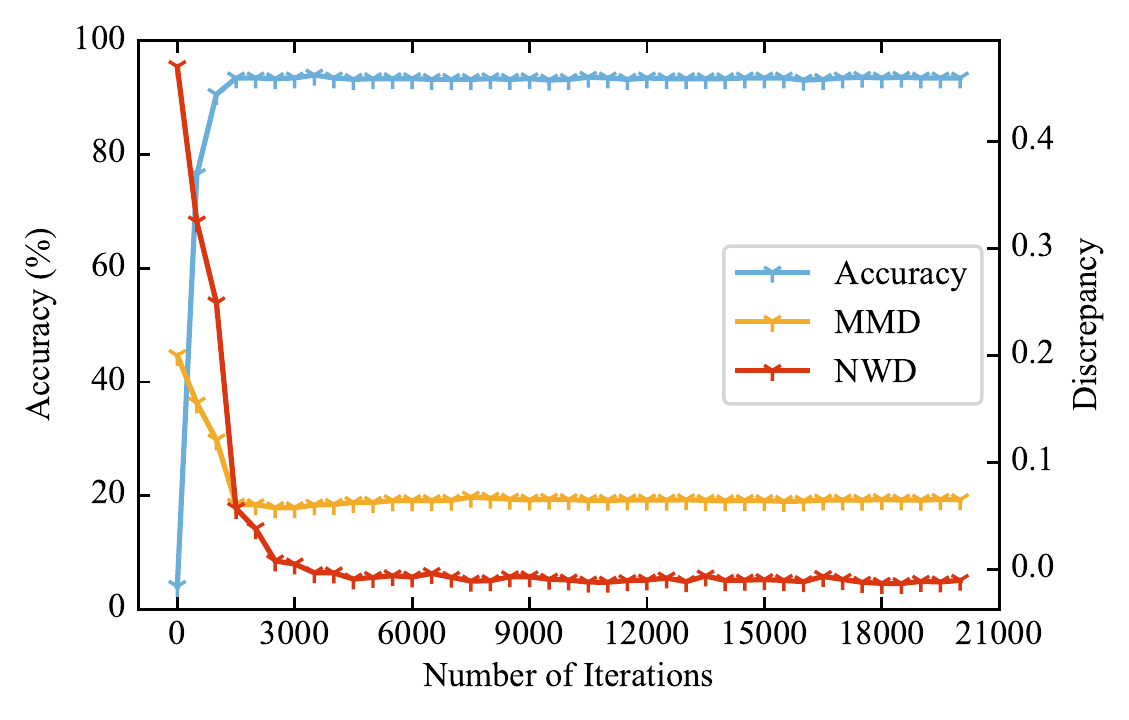}\hfill
}
\subfloat[W→A]{
    \includegraphics[width=0.48\linewidth]{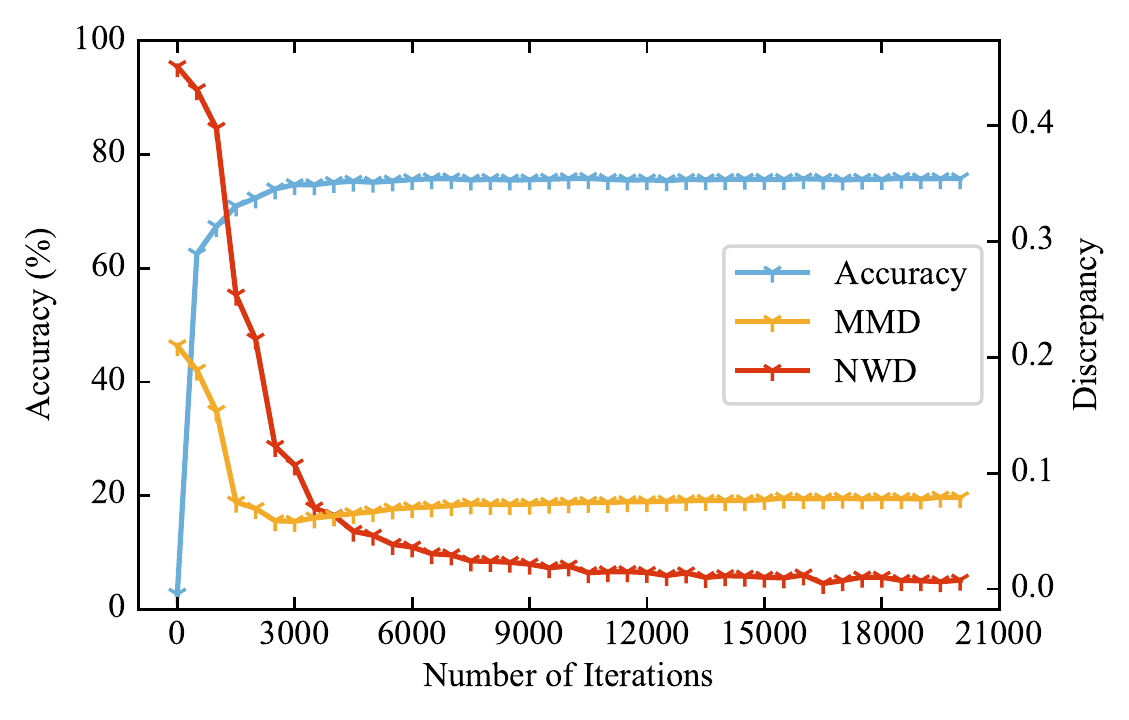}\hfill
}
\vspace{-1mm}
\caption{
The test accuracy, NWD and MMD convergence curves of the target domain on tasks A→W and W→A of Office-31.
}
\vspace{-4mm}
\label{fig:convergence}
\end{figure}

\noindent\textbf{Convergence.} We present the convergence curves of the test accuracy, NWD, and MMD with respect to the number of iterations on tasks A→W and W→A of Office-31, as shown in \Fig{convergence}. Benefiting from the definite guidance meaning, DALN achieves rapid convergence with competitive accuracy. In particular, it can be observed that minimizing the NWD can also effectively decrease the widely-used maximum mean discrepancy (MMD), which also demonstrates the effectiveness of the NWD.

\begin{figure}[htbp]
\centering
\subfloat[Influence of $\lambda$]{\includegraphics[width=0.48\linewidth]{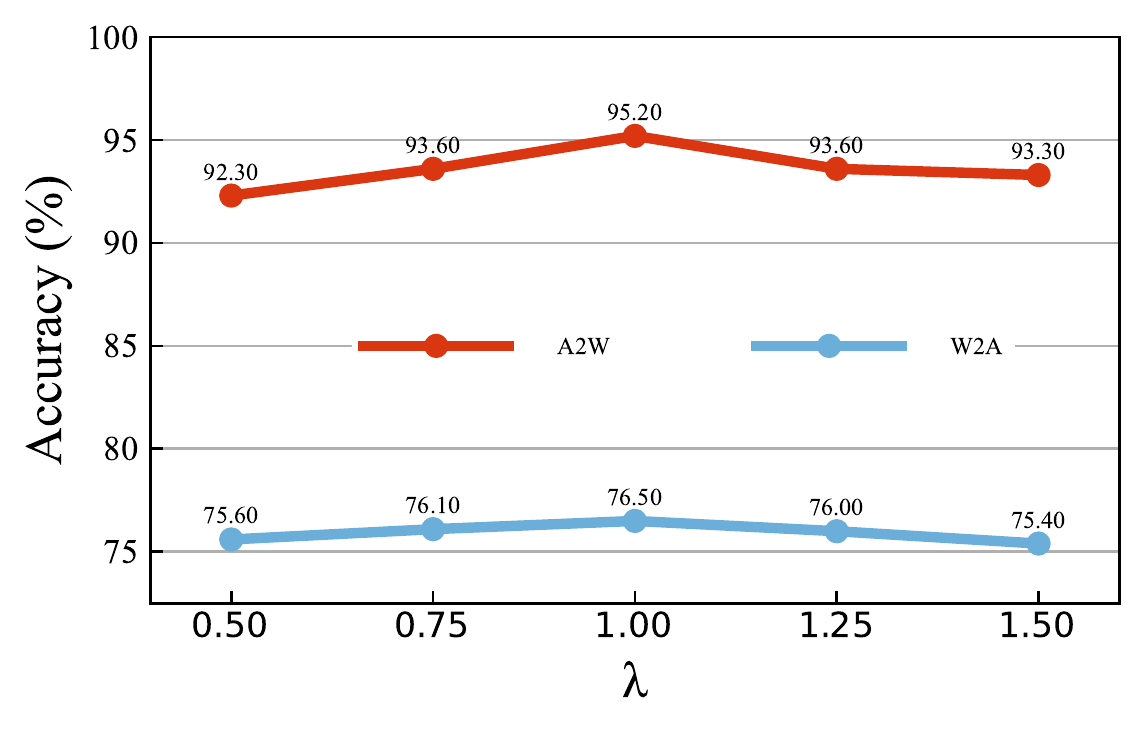}}
\subfloat[Influence of $\gamma$]{\includegraphics[width=0.48\linewidth]{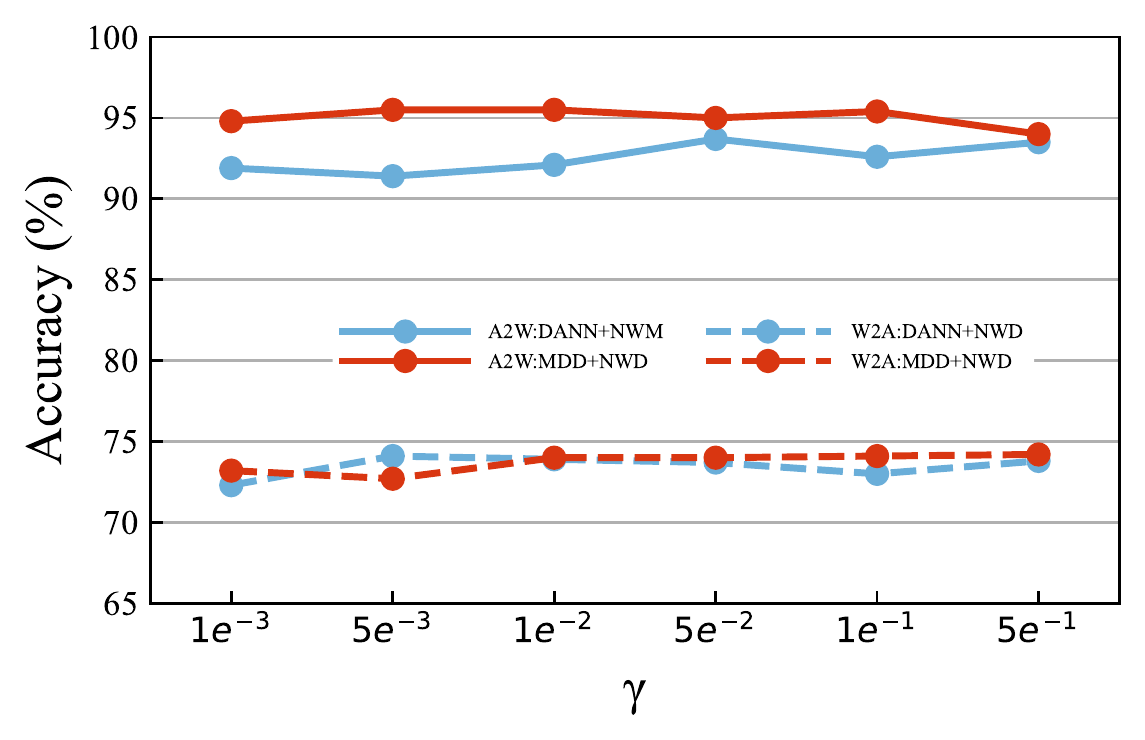}}
\caption{
The influence of $\lambda$ and $\gamma$ on tasks A→W and W→A of Office-31.
}
\label{fig:sensibility}
\end{figure}
\noindent\textbf{Discussion of trade-off parameters $\lambda$ and $\gamma$.} $\lambda$ is used to balance the losses ${\cal L}_{cls}$ and ${\cal L}_{nwd}$. $\gamma$ is also a balance weight used for taking the proposed NWD as a regularizer. Here, we conduct influence analysis for these two parameters based on tasks A→W and W→A of Office-31. As shown in \Fig{sensibility}, DALN achieves the best performance when $\lambda$ ranges from 0.75 to 1.25. For the parameter $\gamma$, pleasing results occur when $\gamma$ is in the range of 0.005 to 0.01. Similar trends can also be  observed in other datasets.  For simplicity, in this work, we set $\lambda$ to 1 and $\gamma$ to 0.01 for all the experiments.

\section{Limitations}
Despite the simplicity and the impressive performance of our method, here comes two problems in the training process. One problem is that the SVD process takes some time to compute the Nuclear norm, and the other problem is that the performance always reaches the highest value early in the training process and then decreases slowly. These two problems will be explored in our future work.



\end{document}